\documentclass[11pt]{article}
\usepackage[english]{babel}
\usepackage[T1]{fontenc}
\usepackage[utf8]{inputenc}
\usepackage{lmodern}
\usepackage[a4paper,left=1.2in,right=1.2in,top=1.1in,bottom=1.1in,footskip=0.5in]{geometry}
\usepackage{amsthm}
\usepackage{amsmath,amsfonts,amssymb}
\usepackage{graphicx}
\usepackage{caption}
\usepackage{subcaption}
\usepackage{authblk} %For author and affiliation
\usepackage{url}            % simple URL typesetting
\usepackage{booktabs}       % professional-quality tables
\usepackage{nicefrac}       % compact symbols for 1/2, etc.
\usepackage{microtype}      % microtypography

\usepackage{enumerate,xspace}
\usepackage{hyperref}
\hypersetup{linkcolor=blue, citecolor=blue, urlcolor=blue, colorlinks=true}

\usepackage{lslbasicmath}
\usepackage{lslcomm}

\usepackage{psfig,epsf}

\RequirePackage{comment}
\theoremstyle{plain}
\newtheorem{theorem}{Theorem}
\newtheorem{lemma}{Lemma}
\newtheorem{proposition}{Proposition}

\newtheorem*{theorem*}{Theorem}
\newtheorem*{lemma*}{Lemma}
\newtheorem*{proposition*}{Proposition}
\newtheorem*{conjecture*}{Conjecture}
\newtheorem{fact*}{Fact}

\theoremstyle{definition}

\newtheorem{example}{Example}
\newtheorem{remark}{Remark}

\newtheorem*{definition*}{Definition}
\newtheorem*{question*}{Question}
\newtheorem*{example*}{Example}
\newtheorem*{remark*}{Remark}
\newtheorem*{remarks*}{Remarks}
\newtheorem*{exercise*}{Exercise}
\newtheorem*{assumption*}{Assumption}

\usepackage[ruled, vlined]{algorithm2e}
\SetKwInput{KwInput}{Input}
\SetKwInput{KwOutput}{Output}
\SetKwInput{KwStart}{Initialization step}
\SetKwInput{KwMinStep}{Minimization step}
\SetKwInput{KwEnd}{Final step}
\SetKwInput{KwInitializationStep}{Initialize}
\SetKwInput{KwUpdateStep}{Update}
\SetKwInput{KwDynUpdateStep}{Dynamic update}
\SetKwInput{KwProcessStep}{Process}
\SetKwInput{KwReturn}{Return}
\SetKwInput{KwGlobalStep}{Global step}
\SetKwInput{KwLocalStep}{Local improvement}

\newcommand{\T}{\mathbf{T}}
\newcommand{\pin}{p_{\rm in}}
\newcommand{\pout}{p_{\rm out}}
\newcommand{\muin}{\mu_{\rm in}}
\newcommand{\muout}{\mu_{\rm out}}

\newcommand{\rin}{r_{\rm in}}
\newcommand{\rout}{r_{\rm out}}
\newcommand{\qin}{q_{\rm in}}
\newcommand{\qout}{q_{\rm out}}
\newcommand{\sout}{s_{\rm out}}
\newcommand{\Fin}{F_{\rm in}}
\newcommand{\Fout}{F_{\rm out}}
\newcommand{\hFin}{\widehat{F}_{\rm in}}
\newcommand{\hFout}{\widehat{F}_{\rm out}}
\newcommand{\hF}{\widehat{F}}
\newcommand{\hG}{\widehat{G}}
\newcommand{\hsigma}{\widehat{\sigma}}

\DeclareMathOperator*{\argmin}{argmin}
\DeclareMathOperator*{\argmax}{argmax}
\DeclareMathOperator*{\Leb}{Leb}

\DeclareMathOperator*{\e}{e}

\DeclareMathOperator*{\sign}{sign}
\newcommand{\rvline}{\hspace*{-\arraycolsep}\vline\hspace*{-\arraycolsep}}

\newcommand{\dHam}{{\rm d_{\rm Ham}}}

\newcommand{\sinc}{\operatorname{sinc}}

\usepackage{times}

\usepackage{import}

\begin{document}

\title{Higher-Order Spectral Clustering for Geometric Graphs}

\author{Konstantin Avrachenkov\footnote{ \texttt{k.avrachenkov@inria.fr}}}
\author{Andrei Bobu\footnote{ \texttt{a.v.bobu@gmail.com}} }
\author{Maximilien Dreveton\footnote{ \texttt{maximilien.dreveton@inria.fr} \\ 
This is the author's version of the article accepted in Springer 
{\it Journal of Fourier Analysis and Applications} 27:22, 2021,
DOI: 10.1007/s00041-021-09825-2 .} }

\affil{Inria Sophia Antipolis, France}
\renewcommand\Authands{ and }

\date{}	
	
\maketitle

\begin{abstract}
The present paper is devoted to clustering geometric graphs. While the standard spectral clustering is often not effective for geometric graphs, we present an effective generalization, which we call higher-order spectral clustering. It resembles in concept the classical spectral clustering method but uses for partitioning the eigenvector associated with a higher-order eigenvalue. We establish the weak consistency of this algorithm for a wide class of geometric graphs which we call Soft Geometric Block Model. A small adjustment of the algorithm provides strong consistency. We also show that our method is effective in numerical experiments even for graphs of modest size.

\bigskip

\end{abstract}

\section{Introduction}
\label{intro}

Graph clustering---the task of identifying groups of tightly connected nodes in a graph---is a widely studied unsupervised learning problem, with applications in computer science, statistics, biology, economy or social sciences \cite{Fortunato_2010}.

In particular, spectral clustering is one of the key graph clustering methods~\cite{von_Luxburg_2007}. In its most basic form, this algorithm consists in partitioning a graph into two communities using the eigenvector associated with the second smallest eigenvalue of the graph's Laplacian matrix (the so-called Fiedler vector~\cite{Fiedler_75}).
Spectral clustering is popular, as it is an efficient relaxation of the NP-hard problem of cutting the graph into two balanced clusters so that the weight between the two clusters is minimal \cite{von_Luxburg_2007}.

In particular, spectral clustering is consistent in the Stochastic Block Model (SBM) for a large set of parameters \cite{abbe_2017}, \cite{lei_rinaldo_2015}.
The SBM is a natural basic model with community structure. It is also the most studied one \cite{abbe_2017}. In this model each node is assigned to one cluster, and edges between node pairs are drawn independently and with probability depending only on the community assignment of the edge endpoints.

However, in many situations, nodes also have geometric attributes (a position in a metric space). Thus, the interaction between a pair of nodes depends not only on the community labelling, but also on the distance between the two nodes.
We can model this by assigning to each node a position, chosen in a metric space.
Then, the probability of an edge appearance between two nodes will depend both on the community labelling and on the positions of these nodes.
Recent proposals of random geometric graphs with community structure include the Geometric Block Model (GBM)~\cite{Galhotra_al_2018} and Euclidean random geometric graphs~\cite{Abbe_Baccelli_Sankararaman_2017}.
The nodes' interactions in geometric models are no longer independent: two interacting nodes are likely to have many common neighbors. While this is more realistic (`friends of my friends are my friends'), this also renders the theoretical study more challenging.

Albeit spectral clustering was shown to be consistent in some specific geometric graphs \cite{Spielman_Teng_2007}, the geometric structure can also heavily handicap a cut-based approach.
Indeed, one could partition space into regions such that nodes between two different regions interact very sparsely.
Thus, the Fiedler vector of a geometric graph might be associated only with a geometric configuration, and bear no information about the latent community labelling.
Moreover, the common technique of regularization~\cite{Zhang_Rohe_2018}, which aims to penalize small size communities in order to bring back the vector associated with the community structure in the second position, will not work in geometric graphs as the regions of space can contain a balanced number of nodes.
Nonetheless, this observation does not automatically render spectral clustering useless.
Indeed, as we shall see, in some situations there is still one eigenvector associated with the community labelling.
Thus, it is now necessary to distinguish the eigenvectors corresponding to a geometric cut---hence potentially useless for cluster recovery---from the one corresponding to the community labelling.
In other words, to achieve a good performance with spectral clustering in such a setting, one needs to select carefully the correct eigenvector, which may no longer be associated with the second smallest eigenvalue.

Our working model of geometric graphs with clustering structure will be the Soft Geometric Block Model (SGBM). It is a block generalization of soft random geometric graphs and includes as particular cases the SBM and the GBM. Another important example is the Waxman Block Model (WBM) where the edge probabilities decrease exponentially with the distance.
The SGBM is similar to the model of \cite{Abbe_Baccelli_Sankararaman_2017}, but importantly we do not assume the knowledge of nodes' positions.

In this paper, we propose a generalization of standard spectral clustering based on a higher-order eigenvector of the adjacency matrix.
This eigenvector is selected using the average intra- and inter-community degrees, and is not necessarily the Fiedler vector. The goal of the present work is to show that this algorithm performs well both theoretically and practically on SGBM graphs.

Our specific contributions are as follows.
We establish weak consistency of higher-order spectral clustering on the SGBM in the dense regime, where the average degrees are proportional to the number of nodes. With a simple additional step, we also establish strong consistency.
One important ingredient of the proof is the characterization of the spectrum of the clustered geometric graphs, and can be of independent interest. In particular, it shows that the limiting spectral measure can be expressed in terms of the Fourier transform of the connectivity probability functions.
Additionally, our numerical simulations show that our method is effective and efficient even for graphs of modest size. Besides, we also illustrate by a numerical example the unsuitability of the Fiedler vector for community recovery in some situations.

Let us describe the structure of the paper.
We introduce in Section~\ref{section:model_definition} the Soft Geometric Block Model and the main notations.
The characterization of the limiting spectrum is given in Section~\ref{section:analysis_limiting_spectrum}.
This characterization will be used in Section~\ref{section:consistency_hosc} to establish the consistency of higher-order spectral clustering in dense SGBM graphs. Finally, Section~\ref{section:numerical_results} shows numerical results and Section~\ref{section:conclusion} concludes the paper with a number of interesting future research directions.

\section{Model definition and notations}
\label{section:model_definition}

\subsection{Notations}

Let $\T^d = \R^d / \Z^d$ be the flat unit torus in dimension $d$ represented by $\left[-\frac12, \frac12 \right]^d$. The norm $\ell_{\infty}$ in $\mathbb R^d$ naturally induces a norm on $\T^d$ such that for any vector $x = (x_1, \ldots, x_d) \in \T^d$ we have $\| x \| = \max\limits_{1 \leq i \leq d} |x_i|$.

For a measurable function $F : \T^d \to \R$ and $k \in \Z^d$, we denote $\hF (k) = \int_{\T^d} F(x) e^{-2i\pi \langle k,x \rangle } \, dx$ the Fourier transform of $F$. The Fourier series of $F$ is given by
\[
\sum_{k \in \Z^d} \hF(k) e^{2i\pi \langle k,x \rangle }.
\]

For two integrable functions $F, \, G: \T^d \to \R$, we define the convolution operation $F * G (y) \weq \int_{\T^d} F(y-x) G(x) \, d x$ and $F^{*m} = F * F * \dots * F$ ($m$ times).
We recall that $\widehat{F * G}(k) = \hF(k) \hG(k)$.

\subsection{Soft Geometric Block Model}

A Soft Geometric Block Model (SGBM) is defined by a dimension $d$, a number of nodes $n$, a set of blocks $K$ and a connectivity probability function $F : \T^d \times K \times K \to [0, 1]$. The node set is taken as $V = [n]$. The model is parametrized by a \textit{node labelling} $\sigma : V \to K$ and \textit{nodes' positions}
$X = (X_1, \dots, X_n) \in \left( \T^d \right)^n$.
We suppose that $F(\cdot, \sigma, \sigma') = F(\cdot, \sigma', \sigma)$ and for any $X \in \T^d$, $F(X)$ depends only on the norm $\|X\|$. The probability of appearance of an edge between nodes $i$ and $j$ is defined by $F\left(X_i-X_j, \sigma_i, \sigma_j \right)$. Note that this probability depends only on the distance between $X_i$ and $X_j$ and the labels $\sigma_i, \sigma_j$.
Consequently, the model parameters specify the distribution
\begin{align}
\label{eq:def_euclidean_block_model}
\pr_{\sigma, X} (A) \weq \prod_{1 \leq i < j \leq n}
\left( F \left( X_i - X_j, \sigma_i, \sigma_j \right) \right)^{A_{ij} }
\left( 1 - F \left( X_i - X_j , \sigma_i, \sigma_j \right) \right)^{1 - A_{ij} }
\end{align}
of the adjacency matrix $A = (A_{ij})$ of a random graph.

Furthermore, for this work we assume that the model has only two equal size blocks, i.\,e., $K = \{1, 2\}$, and $\sum_{i = 1}^n 1(\sigma_i = 1) = \sum_{i = 1}^n 1(\sigma_i = 2) = \frac{n}{2}$. The labels are assigned randomly, that is, the set $\{i \in [n] \colon \sigma_i = 1\}$ is chosen randomly over all the $\frac{n}{2}$-subsets of $[n]$. We assume that the entries of $X$ and $\sigma$ are independent and $\forall i \in V$, $X_i$ is uniformly distributed over $\T^d$. Finally, suppose that for any $x \in \T^d$
\begin{align}
\label{eq:def_homogeneous_model}
  F(x, \sigma, \sigma') \weq
		\left\{
          \begin{array}{ll}
            \Fin(x), & \qquad \mathrm{if}\quad \sigma = \sigma', \\
            \Fout(x), & \qquad \mathrm{otherwise}, \\
          \end{array}
        \right.
\end{align}
where $\Fin, \Fout : \T^d \to [0,1]$ are two measurable functions. We call these functions {\it connectivity probability functions}.

The average intra- and inter-community {\it edge densities} are denoted by $\muin$ and $\muout$, respectively. Their expressions are given by the first Fourier modes of $\Fin$ and $\Fout$:
\begin{align*}
    \muin  \weq \int_{\T^d} \Fin(x) dx \quad \text{ and } \quad  \muout  \weq \int_{\T^d} \Fout(x) dx.
\end{align*}
These quantities will play an important role in the following, as they represent the intensities of interactions between nodes in the same community and nodes in different communities.
In particular, the average inside community degree is $\left(\frac{n}{2}-1 \right) \muin$, and the average outside community degree is $\frac{n}{2}\muout$.

\begin{example}
An SGBM where $\Fin(x) = \pin$ and $\Fout(x) = \pout$ with $\pin, \pout$ being constants is an instance of the Stochastic Block Model.
\end{example}

\begin{example}\label{example:GBM-d_dimension}
An SGBM where $\Fin (x) = 1(\|x\| \leq \rin)$, $\Fout(x) = 1(\|x\| \leq \rout)$ with $\rin, \rout \in \R_+$ is an instance of the Geometric Block Model introduced in \cite{Galhotra_al_2018}.
\end{example}

\begin{example}\label{example:Waxman_Block_Model}
We call Waxman Block Model (WBM) an SGBM with $\Fin (x) = \min( 1,  \qin \e^{-s_{\rm in} || x || })$, $\Fout (x) = \min(1,  \qout \e^{-\sout || x || } )$.
This is a clustered version of the Waxman model \cite{Waxman_1988}, which is a particular case of soft geometric random graphs~\cite{Penrose_2016}.
\end{example}

Formally, {\it clustering} or {\it community recovery problem} is the following problem: given the observation of the adjacency matrix $A$ and the knowledge of $\Fin, \Fout$, we want to recover the latent community labelling $\sigma$.
Given an estimator $\hsigma$ of $\sigma$, we define the loss $\ell$ as the ratio of misclassified nodes, up to a global permutation $\pi$ of the labels:
$
    \ell\left( \sigma, \hsigma \right) \weq \frac{1}{n} \min_{\pi \in \cS_2} \sum_{i} 1\left( \sigma_i \not= \pi \circ \hsigma_i \right).
$
Then, $\hsigma$ is said to be weakly consistent (or equivalently, achieves almost exact recovery) if
\[
\forall \epsilon > 0 \ \colon \ \lim_{n \to \infty} \pr \left( \ell\left( \sigma, \hsigma \right) > \epsilon \right) \weq 0,
\]
and strongly consistent (equivalently, achieves exact recovery) if
\[
\lim_{n \to \infty} \pr \left( \ell\left( \sigma, \hsigma \right) > 0 \right) \weq 0.
\]

\section{The analysis of limiting spectrum}
\label{section:analysis_limiting_spectrum}

\subsection{Limit of the spectral measure}

\begin{theorem}\label{thm:eigenspectrum_adjacency_matrix}
    Consider an SGBM defined by \eqref{eq:def_euclidean_block_model}-\eqref{eq:def_homogeneous_model}.
    Assume that $\Fin(0), \Fout(0)$ are equal to the Fourier series of $\Fin(\cdot),\Fout(\cdot)$ evaluated at~$0$.
	Let $\lambda_1, \dots, \lambda_n$ be the eigenvalues of $A$, and
	\[
	\mu_n \weq \sum_{i=1}^n \delta_{ \lambda_i / n }
	\]
	be the spectral measure of the matrix $\frac{1}{n}A$.
	Then, for all Borel sets $B$ with $\mu \left( \partial B \right) = 0$ and $0 \not\in \bar{B}$, a.s.,
	\[
	\lim_{n \to \infty} \mu_n(B) = \mu(B),
	\]
	where $\mu$ is the following measure:
	\[
	\mu \weq \sum_{k \in \Z^d} \delta_{ \frac{\hFin(k) + \hFout(k)}{2}} \ + \ \delta_{ \frac{\hFin(k) - \hFout(k)}{2}}.
	\]
\end{theorem}

    \begin{remark}\label{remark:limiting_spectrum_explanation}
        The limiting measure $\mu$ is composed of two terms. The first term, $\sum_{k \in \Z^d} \delta_{ \frac{\hFin(k) + \hFout(k)}{2}}$ corresponds to the spectrum of a random graph with no community structure, and where edges between two nodes at distance $x$ are drawn with probability $\frac{\Fin(x) + \Fout(x)}{2}$. In other words, it is the null-model of the considered SGBM. Hence, the eigenvectors associated with those eigenvalues bear no community information, but only geometric features.

        On the contrary, the second term $\sum_{k \in \Z^d} \delta_{ \frac{\hFin(k) - \hFout(k)}{2}}$ corresponds to the difference between intra- and inter-community edges. In particular, as we shall see later, the ideal eigenvector for clustering is associated with the eigenvalue $\widetilde \lambda$ closest to $ \lambda_* = n \frac{\hFin(0) - \hFout(0)}{2} $. Other eigenvectors might mix some geometric and community features and hence are harder to analyze.

        Last, the eigenvalue $ \widetilde \lambda $ is not necessarily the second largest eigenvalue, as the ordering of eigenvalues here depends on the Fourier coefficients $\hFin(k)$ and $\hFout(k)$, and is in general non trivial.
    \end{remark}

    \begin{remark}
        The assumptions on $\Fin(0)$ and $\Fout(0)$ are validated for a wide range of reasonable connectivity functions. For instance, by Dini's criterion, all the functions that are differentiable at 0 satisfy these conditions. Another appropriate class consists of piecewise $C^1$ functions that are continuous at 0 (this follows from the Dirichlet conditions).
    \end{remark}

    \begin{proof}
    The outline of the proof of Theorem~\ref{thm:eigenspectrum_adjacency_matrix} follows closely~\cite{Bordenave_2008}.
	First, we show that $\forall m \in \N$, $\lim\limits_{n \to \infty} \, \E \, \mu_n \left( P_m \right) = \mu (P_m)$ where $P_m(t) = t^m$.
	Second, we use Talagrand's concentration inequality to prove that $\mu_n(P_m)$ is not far from its mean, and conclude with Borel--Cantelli lemma.

	(i) By Lemma \ref{lemma:moment_problem_mu} in the Appendix, in order to establish the desired convergence it is enough to show that $\lim\limits_{n \to \infty} \E \mu_n \left( P_m \right) = \mu (P_m)$ for any $m\in\N$.
	First,
	\begin{align}\label{in_proof_eq:link_measure_moments}
	   \E \mu_n(P_m) \weq \dfrac{1}{n^m} \sum_{i=1}^n \E \lambda_i^m \weq\dfrac{1}{n^m}\E \Tr A^m.
	\end{align}
	By definition,
	\begin{align*}
	\Tr A^m & \weq \sum_{\alpha \in [n]^m } \prod_{j=1}^m A_{i_j, i_{j+1} },
	\end{align*}
	with $\alpha = (i_1, \dots, i_m) \in [n]^m $ and $i_{m+1} = i_1$. We denote $\cA^m_n$ the set of $m$-permutations of $[n]$, that is $\alpha \in \cA^m_n$ iff $\alpha$ is an $m$-tuple without repetition.
	We have,
	\begin{align}\label{in_proof_moments}
	\Tr A^m & \weq \sum_{\alpha \in \cA^m_n } \, \prod_{j=1}^m A_{i_j, i_{j+1} } + R_m,
	\end{align}
	where $R_m = \sum\limits_{ \alpha \in [n]^m \backslash \cA^m_n } \, \prod_{j=1}^m A_{i_j, i_{j+1} } $.
	
	We first bound the quantity $R_m$. Since $| A_{ij}| \leq 1$, we have
	\begin{align*}
	|R_m| &\wle \ \Big| [n]^m \backslash \cA^m_n \Big| \weq n^m - \dfrac{n!}{(n-m)! } \weq \frac{m(m-1)}{2}  n^{m-1}  + o(n^{m-1}),
	\end{align*}
	where we used $\dfrac{n!}{(n-m)! }  = n^m - n^{m-1} \sum_{i=0}^{m-1} i + o(n^{m-1}) $.
	Hence
	\begin{align}\label{in_proof_remaining_term_neglected}
	\lim\limits_{n \to \infty} \dfrac{1}{n^m} R_m = 0.
	\end{align}
	Moreover,
	\begin{align*}
	\E \sum_{\alpha \in \cA^m_n } \, \prod_{j=1}^m A_{i_j, i_{j+1} }
	& \weq \sum_{\alpha \in \cA^m_n } \, \int_{(\T^d ) ^m } \prod_{j=1}^m F( x_{i_j} - x_{i_{j+1}}, \sigma_{i_j}, \sigma_{i_{j+1} }  ) dx_{i_1} \dots dx_{i_m} \\
	& \weq \sum_{\alpha \in \cA^m_n } \, G(\alpha)
	\end{align*}
	where $G(\alpha) = \int_{(\T^d ) ^m } \prod_{j=1}^m F( x_{i_j} - x_{i_{j+1}}, \sigma_{i_j}, \sigma_{i_{j+1} } ) dx_{i_1} \dots dx_{i_m} $ for $\alpha \in \cA^m_n$.
	
	Let us first show that the value of $G(\alpha)$ depends only on the number of consecutive indices corresponding to the nodes from the same community.
	More precisely, let us define the set $\mathcal S(\alpha) = \{ j \in [m] : \sigma_{i_j} = \sigma_{i_{j+1} }  \}$. Using Lemma~\ref{appendix:lemma_convolution_m_times} in the Appendix and the fact that the convolution is commutative, we have
	\[
	    G(\alpha) \weq \Fin^{* |\mathcal S(\alpha)|} * \Fout^{* (m-|\mathcal S(\alpha)|)} (0).
	\]
	We introduce the following equivalence relationship in $\cA^m_n$: $\alpha \sim \alpha'$ if $|\mathcal S(\alpha)| = |\mathcal S(\alpha')|$.
	We notice than $G(\cdot)$ is constant on each equivalence class, and equals to $\Fin^{* p} * \Fout^{* (m-p)} (0)$ for any $\alpha \in \cA^m_n $ such that $|\mathcal S(\alpha)| = p$.
	%Since $G(\cdot)$ is constant on each equivalence class, we denote by $G_p$ the value of $G(\alpha)$ taken for any $\alpha \in \cA^m_n $ such that $|\mathcal S(\alpha)| = p$.
	
	Then, let us calculate the cardinal of each equivalence class with $|\mathcal S(\alpha)| = p$. First of all, we choose the set $\mathcal S(\alpha)$ which can be done in $\binom mp$ ways if $m-p$ is even and cannot be done if $m-p$ is odd. The latter follows from the fact that $p$ (the number of `non-changes' in the consecutive community labels) has the same parity as $m$ (the total number of indices) since $i_{m+1} = i_1$. The set $\mathcal S(\alpha)$ defines the community labels up to the flip of communities since $\sigma_{i_j} = \sigma_{i_{j+1}}$ for any $j \in \mathcal S(\alpha)$ and $\sigma_{i_j} \neq \sigma_{i_{j+1}}$ for $j \in [m] \backslash \mathcal S(\alpha)$.
	
	Let $N_1(\alpha)$ be the number of indices $i_j$ with $\sigma_{i_j} = 1$. Consider first the case $\sigma_{i_1} = 1$ and note that $N_1(\alpha)$ is totally defined by the set $\mathcal S(\alpha)$. There are $\frac{n}{2}$ possible choices for $i_1$. Now we have two possibilities. If $\sigma_{i_1} = \sigma_{i_2}$ then we have $\frac{n}{2}-1$ possible choices for the index $i_2$ (since $\alpha \in \mathcal A_n^m$). Otherwise, if $\sigma_{i_1} \neq \sigma_{i_2}$ then the index $i_2$ can be chosen in $\frac{n}{2}$ ways. Resuming the above operation, we choose $N_1(\alpha)$ indices from the first community, and it can be done in $n/2(n/2 - 1)\ldots (n/2 - N_1(\alpha)) $ ways. The indices from the second community can be chosen in $n/2(n/2 - 1) \ldots (n/2 - (m - N_1(\alpha)))$ ways. Thus in total the number of possible choices of indices is
	\begin{align*}
	    & \frac n2 \left(\frac n2 - 1\right)\ldots \left(\frac n2 - N_1(\alpha)\right) \cdot \frac n2\left(\frac n2 - 1\right) \ldots \left(\frac n2 - (m - N_1(\alpha))\right)  \\
	    &= \frac{n^m}{2^m} + O(n^{m-1}), \;\; n \to\infty.
	\end{align*}
	The same reasoning applies if $\sigma_{i_1} =2$.
	Hence, when $n$ goes to infinity, the cardinal of each equivalence class is
	\[
	\left| \{ \alpha \in \mathcal A_n^m \colon |\mathcal S(\alpha)| = p\} \right| \weq
	\begin{cases}
	0 & \text{ if } m-p \text{ is odd,}\\
	2 \binom{m}{p} \frac{n^m}{2^m} + O(n^{m-1}) & \text{ otherwise.}
	\end{cases}
	\]
	This can be rewritten as
	$$
	    \left| \{ \alpha \in \mathcal A_n^m \colon |\mathcal S(\alpha)| = p\} \right| = \binom{m}{p} \left( 1 + (-1)^{m-p}  \right) \frac{n^m}{2^m} + O(n^{m-1}), \;\; n \to\infty.
	$$
    Hence,
	\begin{align*}
	\E \sum_{\alpha \in \cA^m_n } \, \prod_{j=1}^m A_{i_j, i_{j+1} } & \weq \sum_{p=0}^{n} \left| \{ \alpha \in \mathcal A_n^m \colon |\mathcal S(\alpha)| = p\} \right| \Fin^{* p} * \Fout^{* (m-p)} (0)
	 \\
	%& \weq \frac{n^m}{2^m} \ \sum_{p = 0}^m \binom{m}{p} \left( 1 + (-1)^{m-p}  \right) G_p + O(n^{m-1}) \\
	& \weq \frac{n^m}{2^m} \ \sum_{p = 0}^m \binom{m}{p} \left( 1 + (-1)^{m-p} \right)  \Fin^{* p}* \Fout^{* (m-p)} (0) + O(n^{m-1}) \\
	%& \weq \textcolor{blue}{ \frac{n^m}{2^m} \left( \sum_{p = 0}^m \binom{m}{p} \Fin^{* p}* \Fout^{* (m-p)} (0) \ + \ \sum_{p = 0}^m \binom{m}{p} \Fin^{* p}* (-\Fout)^{* (m-p)} (0) \right)	+ O(n^{m-1}) } \\
	& \weq n^m \left( \left(\frac{ \Fin + \Fout }{2} \right)^{*m} (0)   + \left( \frac{ \Fin - \Fout }{2} \right)^{*m} (0) \right) + O(n^{m-1}).
	\end{align*}
	Therefore, equations~\eqref{in_proof_eq:link_measure_moments}, \eqref{in_proof_moments} and~\eqref{in_proof_remaining_term_neglected} give us:
	\begin{align*}
	\lim_{ n\to \infty } \E \mu_n (P_m) \weq \left( \frac{ \Fin + \Fout }{2} \right)^{*m} (0)
	+ \left( \frac{ \Fin - \Fout }{2} \right)^{*m} (0).
	\end{align*}
	
	Finally, since $\Fin, \Fout$ are equal to their Fourier series at~$0$, and using $ \widehat{F * G} (k) = \hF(k) \hG(k)$, we have
	\begin{equation}\label{eq:lim_E_mu_n}
	\lim_{ n\to \infty } \E \mu_n (P_m) \weq \sum_{k \in \Z^d} \left(  \frac{ \hFin (k) + \hFout (k) }{2}  \right) ^ m
	+ \left(  \frac{ \hFin (k) - \hFout (k) }{2}   \right) ^ m  = \mu \left( P_m \right).
	\end{equation}

	(ii) For each $m \geq 1$, and $n$ fixed, we define
	\begin{align*}
	     \begin{array}{lcl}
            Q_m : & SGBM(\Fin, \Fout) %\{0,1\}^{n\times n}
            & \longrightarrow \R \\
                & A & \longmapsto \frac{1}{n^{m-1}} \Tr A^{m} \\
          \end{array}
	\end{align*}
	where $SGBM(\Fin, \Fout)$ denotes the set of adjacency matrices of an SGBM random graph with connectivity functions $\Fin, \Fout$.
	Note that $Q_m(A) = n \mu_n(P_m)$.
	
	Let $A, \widetilde{A}$ be two adjacency matrices. We denote the Hamming distance by $\dHam \left( A, \widetilde{A} \right) = \sum_{i=1}^n \sum_{j=1}^n 1( A_{ij} \not= \widetilde{A}_{ij})$.
	Using Lemma~\ref{lemma:appendix_Tr_1_lischitz} in the Appendix, we show that the function $Q_m$ is $(m/n)$--Lipschitz for the Hamming distance:
	\begin{align}
	\label{in_proof_eq_Qm_lipshitz}
	    \left| Q_m (A) - Q_m ( \widetilde{A} ) \right| & \wle  \frac{m}{n} \ \dHam\left( A, \widetilde{A} \right).
	\end{align}
	
	Let $M_m$ be the median of $Q_m$. Talagrand's concentration inequality \cite[Proposition~2.1]{Talagrand_1996} states that
	\begin{align}\label{ineq:in_proof_talagrand}
	    \pr \left( \left| Q_m - M_m \right| > t \right) \wle 4 \exp \left( - \frac{ n^2 t^2 }{ 4 m^2 } \right),
	\end{align}
	which after integrating over all $t$ gives
	\begin{align*}
	    \left| n \E \mu_n \left( P_m \right) - M_m \right| \wle \E \left| Q_m(A) - M_m \right| \wle \frac{C_m}{n},
	\end{align*}
	since $\E X = \int_{0}^\infty \pr(X>t) dt$ for any positive random variable $X$. The constant $C_m$ is equal to $8 m \int_0^\infty e^{-u^2} du$.
	
	Moreover,
	\begin{align*}
	    n \left| \mu_n(P_m) - \E \mu_n(P_m) \right| & \wle \left| n \mu_n(P_m) - M_m \right| + \left| M_m - n \E \mu_n(P_m) \right| \\
	    & \wle \left| Q_m - M_m \right| + \frac{C_m}{n}.
	\end{align*}
	Let $s > 0$. Since $C_m / n^2$ goes to $0$ when $n$ goes to infinity, we can pick $n$ large enough such that $s > \frac{C_m}{n^2}$. Thus, using again inequality~\eqref{ineq:in_proof_talagrand}, we have
	%Thus, using again inequality~\eqref{ineq:in_proof_talagrand}, we have for all $s > \frac{C_m}{n^2}$,
	\begin{align*}
	    \pr \left( \left|  \mu_n(P_m) - \E \mu_n(P_m) \right| > s \right) & \wle \pr \left( \frac{1}{n} \left| Q_m - M_m \right| > s - \frac{C_m}{n^2} \right) \\
	    & \wle 4 \exp \left( - \frac{n^4}{4m^2} \left( s - \frac{C_m}{n^2} \right)^2 \right).
	\end{align*}
	However, by \eqref{eq:lim_E_mu_n}, $\lim\limits_{n \to \infty} \E \mu_n(P_m) = \mu(P_m)$. Hence $\mu_n(P_m)$ converges in probability to $\mu(P_m)$.
	Let $s_n =  \frac{1}{n^\kappa}$ with $\kappa > 0$, and $$
	\epsilon_n = 4 \exp \left( - \frac{n^4}{4m^2} \left( s_n - \frac{C_m}{n^2} \right)^2 \right).
	$$
	Since $ \sum_{n \in \N} \epsilon_n < \infty $ when $\kappa < 2$,
	an application of Borel--Cantelli lemma shows that the convergence holds in fact almost surely.
	This concludes the proof.
    \end{proof}

\subsection{Conditions for the isolation of the ideal eigenvalue}

    As noticed in Remark~\ref{remark:limiting_spectrum_explanation}, the ideal eigenvector for clustering is associated with the eigenvalue of the adjacency matrix $A$ closest to
    $n \frac{\muin - \muout}{2}$ (recall that $\muin = \hFin(0)$ and $\muout = \hFout(0)$). The following proposition shows that this ideal eigenvalue is isolated from other eigenvalues under certain conditions.

    \begin{proposition}\label{proposition:unicity_eigenvector_general_case}
        Consider the adjacency matrix $A$ of an SGBM defined by~\eqref{eq:def_euclidean_block_model}-\eqref{eq:def_homogeneous_model}, and assume that:
    \begin{align}
        \hFin(k) + \hFout(k) & \not= \muin - \muout,  \qquad \forall k \in \Z^d \label{condition:Fourier_mode_different_ideal_mode_meansum}, \\
        \hFin(k) - \hFout(k) & \not= \muin - \muout, \qquad \forall k \in \Z^d \backslash \{0\}, \label{condition:Fourier_mode_different_ideal_mode_meandifference}
    \end{align}
    with $\muin \neq \muout$. Then, the eigenvalue of $A$ closest to $n \frac{\muin - \muout}{2}$ is of multiplicity one. Moreover, there exists $\epsilon > 0$ such that for large enough $n$ every other eigenvalue is at distance at least $\epsilon n$.
    \end{proposition}

    \begin{proof}
        Let $\lambda_1, \dots, \lambda_n$ be the eigenvalues of $A$. Let $i^* \in \argmin_{i \in [n]} \left| \frac{\lambda_i}{n} - \frac{\muin - \muout}2 \right|$.
        We shall show that there exists $\epsilon >0$ such that for large enough $n$, we have for all $i \not= i^*$:
    \begin{align*}
        \left| \frac{\lambda_i}{n} - \frac{\muin - \muout}2 \right| > \epsilon.
    \end{align*}
    Due to  condition~\eqref{condition:Fourier_mode_different_ideal_mode_meansum}, and the fact that
        $$
            \lim_{|k|\to\infty} \left( \hFin(k) + \hFout (k)\right) = 0,
        $$
        there is some fixed $\epsilon_1 > 0$ such that
        $$
            \min_{k \in \Z^d} \left(  \left| \frac{ \hFin(k) + \hFout(k) }{2} - \frac{\muin - \muout}2 \right| \right) > \epsilon_1.
        $$
        Similarly, condition~\eqref{condition:Fourier_mode_different_ideal_mode_meandifference} ensures the existence of $\epsilon_2 >0$ such that
        $$
            \min_{k \in \Z^d \backslash \{0\} } \left( \left| \frac{ \hFin(k) - \hFout(k) }{2} - \frac{\muin - \muout}2 \right| \right) > \epsilon_2.
        $$
        Denote $\epsilon_3 = \frac{|\muin - \muout|}4$. Let $\epsilon = \min\left(\epsilon_1, \epsilon_2, \epsilon_3\right)$, and consider the interval $B = \left[ \frac{\muin - \muout }{2} - \epsilon, \frac{\muin - \muout }{2} + \epsilon \right ]$. By Theorem \ref{thm:eigenspectrum_adjacency_matrix}, a.s.,
        $$
            \lim_{n\to\infty} \mu_n(B) = \mu(B) = 1.
        $$
        Therefore, for $n$ large enough the only eigenvalue of $A$ in the interval $B$ is $\lambda_{i^*}$.
        \end{proof}

The following proposition shows that conditions~\eqref{condition:Fourier_mode_different_ideal_mode_meansum} and~\eqref{condition:Fourier_mode_different_ideal_mode_meandifference} of Proposition~\ref{proposition:unicity_eigenvector_general_case} are almost always verified for a GBM.

\begin{proposition}\label{cor:gap_between_good_eigenvaue_and_others}
    Consider the $d$-dimensional GBM model, where $\Fin, \Fout$ are 1-periodic, and defined on the flat torus $\T^d$ by $\Fin(x) = 1(\|x\| \leq \rin)$ and $\Fout(x) = 1(\|x\| \leq \rout)$, with $\rin > \rout > 0$. Denote by $\mathcal B_+$ and $\mathcal B_-$ the sets of parameters $\rin$ and $\rout$ defined by negation of conditions \eqref{condition:Fourier_mode_different_ideal_mode_meansum} and \eqref{condition:Fourier_mode_different_ideal_mode_meandifference}:
    \begin{align*}
        \mathcal B^+ &= \left\{ (\rin, \rout) \in \mathbb R^2_+ \colon \hFin(k) + \hFout(k) = \muin - \muout \text{ for some } k \in \Z^d \right\} \\
        \mathcal B^- &= \left\{ ( \rin, \rout ) \in \mathbb R^2_+ \colon \hFin(k) - \hFout(k) = \muin - \muout \text{ for some } k \in \Z^d \backslash\{0\} \right\}.
    \end{align*}
    Then these sets of `bad' parameters are of zero Lebesgue measure:
        \begin{align*}
            \mathrm{Leb}(\mathcal B^+) \weq 0; \qquad \text{and} \qquad
            \mathrm{Leb}(\mathcal B^-) \weq 0.
        \end{align*}
    Hence for $\mathcal B = \mathcal B^+ \cup \mathcal B^-$
    \[
        \mathrm{Leb}(\mathcal B) = 0.
    \]
    \end{proposition}

    \begin{proof}

    It is clear that
    $$
        \Leb(\mathcal B) \leq \Leb(\mathcal B^+) + \Leb(\mathcal B^-).
    $$
    Thus, it is enough to show that $\Leb(\mathcal B^+) = 0$ and $\Leb(\mathcal B^-) = 0$. We shall establish the first equality, and the second equality can be proved similarly.

    By Lemma \ref{lemma:computation_fourier_transform_square_signal} in the Appendix, the condition \eqref{condition:Fourier_mode_different_ideal_mode_meansum} for given functions $\Fin$ and $\Fout$ is as follows:
    \begin{align*}
        & \rin^d \prod_{j = 1}^d \sinc(2\pi \rin k_j) + \rout^d \prod_{j = 1}^d \sinc(2\pi \rout k_j) = \\
        &= \rin^d - \rout^d \text{ for some $k = (k_1, \ldots, k_d) \in \mathbb Z^d$}.
    \end{align*}
    Notice that $\lim_{k_j \to \infty} \sinc(2\pi \rin k_j) = 0$ and $\lim_{k_j \to \infty} \sinc(2\pi \rout k_j) = 0$ while the right-hand side of the above equation is fixed. Therefore, this equation can hold only for $k$ from a finite set $\mathcal K$. Let us fix some $k = (k_1, \ldots, k_d) \in \mathcal K$ and denote
    \begin{multline*}
        \mathcal B^+_k = \Biggl\{ (\rin, \rout) \in \mathbb R^2_+ \colon \rin^d \prod_{j = 1}^d \sinc(2\pi \rin k_j) + \rout^d \prod_{j = 1}^d \sinc(2\pi \rout k_j) = \rin^d - \rout^d \Biggr\}.
    \end{multline*}

    Let us now fix $\rin$, and consider the condition defining $\mathcal B_k^+$ as an equation on $\rout$. Define the functions
    \begin{align*}
        f_k(x) &= x^d \left(1 + \prod_{j = 1}^d \sinc(2\pi x k_j)\right); \\
        g_k(x) &= x^d \left(1 - \prod_{j = 1}^d \sinc(2\pi x k_j)\right).
    \end{align*}
    Then the mentioned equation takes the form
    \begin{equation}\label{eq:prop-gbm-main-equation}
        f_k(\rout) = g_k(\rin).
    \end{equation}
        Consider the function $h_k: \mathbb C \to \mathbb R$:
    $$
        h_k(z) = z^d \left(1 + \prod_{j = 1}^d \sinc(2\pi z k_j)\right).
    $$
    Clearly, this function coincides with $f_k$ on $\mathbb R$. Moreover, it is holomorphic in $\mathbb C$, as $\sinc(z)$ is holomorphic (it can be represented by the series $\sum_{n = 0}^{\infty} \frac{(-1)^n}{(2n+1)!} z^{2n}$), and the product of holomorphic functions is again holomorphic. But then the derivative $h'_k(z)$ is also holomorphic, therefore, it has a countable number of zeros in $\mathbb C$. Clearly, $h'_k \equiv f'_k$ on $\mathbb R$, which yields that $f'_k$ has a countable number of zeros in $\mathbb R$.

    Hence, $\mathbb R_+$ is divided into a countable number of intervals on which the function $f_k(x)$ is strictly monotone. That is, $\mathbb R_+ = \sqcup_{i = 0}^{\infty} [a_i(k), b_i(k)]$ where $f_{k, i} = f_{k} \bigr|_{[a_i(k), b_i(k)]}$ is strictly monotone. Then the function $f_{k, i}^{-1}(x)$ is correctly defined and, since $f_{k, i}$ is measurable and injective, $f_{k, i}^{-1}$ is measurable as well. Consequently, there is a unique solution $\rout = f_{k, i}^{-1}(g_k(\rin))$ of equation  \eqref{eq:prop-gbm-main-equation} for $\rin \in [\min f_{k,i}; \max f_{k,i}]$. If $\rin \not\in [\min f_{k,i}; \max f_{k,i}]$, there is no solution at all.

    Therefore, $\mathcal B^+_{k, i} = \left\{ \left(\rin, f_{k, i}^{-1}(g_k(\rin)) \right) \colon \rin \in [\min f_{k,i}; \max f_{k,i}] \right\}$ is the graph of some measurable function in $\mathbb R_+^2$. Since such a graph has a zero Lebesgue measure (see \textit{e.g.}, \cite[Lemma~5.3]{Wheeden_Zygmund_77}), we have:
    $$
        \mathrm{Leb}(\mathcal B^+_k) = \mathrm{Leb}\left(\cup_{i = 1}^{\infty} \mathcal B^+_{k, i}\right) = 0.
    $$
    Hence, we can conclude that
    $$
        \Leb(\mathcal B^+) = \Leb\left(\bigcup_{k \in \mathcal K} \mathcal B^+_k\right) \leq \sum_{k \in \mathcal K} \Leb(\mathcal B^+_k) = 0.
    $$
    Carrying out similar argumentation for $\mathcal B^-$ completes the proof.
\end{proof}

\section{Consistency of higher-order spectral clustering}
\label{section:consistency_hosc}

In this section we show that spectral clustering based on the ideal eigenvector (see Algorithm~\ref{algo:higher_order_sc}) is weakly consistent for SGBM (Theorem~\ref{thm:almost_recovery_sgbm}).
We then show that a simple extra step can in fact achieve strong consistency.

\begin{algorithm}
	\KwInput{Adjacency matrix $A$, average intra- and inter-cluster edge densities $\muin, \muout$.
	}
	
	\KwOutput{Node labelling $\widetilde{\sigma} \in \{1, 2\}^n$.}~\\
	\KwGlobalStep \\
	{   Let $\widetilde{\lambda}$ be the eigenvalue of $A$ closest to $\lambda_* = \frac{\left(\muin-\muout\right)}{2} n$, and $\widetilde{v}$ be the associated eigenvector. \\
	\For{$i=1,\dots,n$}
	{
		If $\widetilde{v}_i >0$, let $\widetilde{\sigma}_i = 1$; otherwise, let $\widetilde{\sigma}_i = 2$.
	}
	}
	\caption{Higher-Order Spectral Clustering (HOSC).}
	\label{algo:higher_order_sc}
\end{algorithm}

\begin{remark}
    The worst case complexity of the eigenvalue factorization is $O\left( n^3 \right)$ \cite{demmel2008}. However, when the matrix is sufficiently sparse and the eigenvalues are well separated, the empirical complexity can be close to $O(kn)$, where $k$ is the number of required eigenvalues \cite{demmel2008}. Moreover, since Algorithm~\ref{algo:higher_order_sc} uses only the sign of eigenvector elements, a quite rough accuracy can be sufficient for classification purposes.
\end{remark}

\subsection{Weak consistency of higher-order spectral clustering}

\begin{theorem}\label{thm:almost_recovery_sgbm}
    Let us consider the $d$-dimensional SGBM with connectivity probability functions $F_{in}$ and $F_{out}$ satisfying conditions~\eqref{condition:Fourier_mode_different_ideal_mode_meansum}-\eqref{condition:Fourier_mode_different_ideal_mode_meandifference}. Then Algorithm~\ref{algo:higher_order_sc} is weakly consistent. More precisely, Algorithm~\ref{algo:higher_order_sc} misclassifies at most $O(\log n)$ nodes.
\end{theorem}

\begin{proof}
Let us introduce some notations.
Recall that $\muin = \hFin(0)$ and $\muout = \hFout(0)$. In the limiting spectrum, the ideal eigenvalue for clustering is
\[
    \lambda_* = \frac{\muin - \muout}{2} n.
\]
We consider the closest eigenvalue of $A$ to $\lambda_*$:
$$
\widetilde\lambda = \argmin\limits_{\lambda}\,(|\lambda - \lambda_*|).
$$
Also, let $\widetilde v$ be the normalized eigenvector corresponding to $\widetilde \lambda$. In this proof, the Euclidean norm $\| \cdot \|_2$ is used.

The plan of the proof is as follows. We consider the vector
$$
    v_* = (\underbrace{1/\sqrt{n}, \ldots, 1/\sqrt n}_{n/2}, \underbrace{-1/\sqrt{n}, \ldots, -1/\sqrt n}_{n/2})^{\mathrm T},
$$
where we supposed without loss of generality that the $n/2$ first nodes are in Cluster~1, and the $n/2$ last nodes are
in Cluster~2.
The vector $v_*$ gives the perfect recovery by the sign of its coordinates. We shall show that with high probability for some constant $C > 0$
\begin{equation}\label{x-approx}
    \| \widetilde v - v_* \|_2 \leq C \sqrt{\frac{\log n}n}.
\end{equation}
We say that an event occurs \textit{with high probability} (\textit{w.\,h.\,p.}) if its probability goes to 1 as $n\to\infty$. With the bounding~\eqref{x-approx}, we shall then show that at most $o(n)$ of entries of $\widetilde v$ have a sign that differs from the sign of the respective entry in $v_*$; hence $\widetilde v$ retrieves almost exact recovery.

In order to establish inequality \eqref{x-approx} we shall use the following theorem from~\cite{kahan_parlett_1982}.
\begin{theorem}\label{th-kanan-parlett} Let $A$ be a real symmetric matrix. If $\widetilde \lambda$ is the eigenvalue of $A$ closest to $\rho(v) = \frac{v^T A v}{v^T v}$, $\delta$ is the separation of $\rho$ from the next closest eigenvalue and $\widetilde v$ is the eigenvector corresponding to $\widetilde \lambda$, then
    $$
        |\sin \angle(v, \widetilde v)| \leq \frac{\| Av - \rho v \|_2}{\|v\|_2 \delta}.
    $$
\end{theorem}

First we deal with $\rho(v_*)$. Since $v_*$ is normalized and real-valued (by the symmetry of $A$), we have
$$
    \rho(v_*) = v^T_* A v_*.
$$
Denote $u = Av_*$. Then, obviously,
\begin{equation}\label{y_i_expression}
    u_i = \sum_{j = 1}^n A_{ij} (v_*)_i = \frac 1{\sqrt n}\sum_{j = 1}^{n/2}{A_{ij}} - \frac 1{\sqrt n}\sum_{j = n/2+1}^{n}{A_{ij}}.
\end{equation}
It is clear that each entry $A_{ij}$ with $i \neq j$ is a Bernoulli random variable with the probability of success either $\muin$ or $\muout$. This can be illustrated by the element-wise expectation of the adjacency matrix:
    $$
        \mathbb EA =
            \begin{pmatrix}
              \begin{matrix}
\muin & \dots & \muin \\
\vdots & \ddots & \vdots \\
\muin & \dots & \muin
\end{matrix} & \rvline & \begin{matrix}
\muout & \dots & \muout \\
\vdots & \ddots & \vdots \\
\muout & \dots & \muout
\end{matrix}  \\
            \hline
            \begin{matrix}
\muout & \dots & \muout \\
\vdots & \ddots & \vdots \\
\muout & \dots & \muout
\end{matrix}
& \rvline &
              \begin{matrix}
\muin & \dots & \muin \\
\vdots & \ddots & \vdots \\
\muin & \dots & \muin
\end{matrix} \\
            \end{pmatrix}.
    $$

Let us consider the first term in the right-hand side of \eqref{y_i_expression} for $i \leq n/2$. Since $A_{ij}$ are independent for fixed $i$, it is easy to see that $Y_i := \sum_{j = 1}^{n/2}{A_{ij}} \sim \mathrm{Bin}(n/2-1, \muin)$ with the expectation $\mathbb EY_i = (n/2-1)\muin$. Then we can use the Chernoff bound to estimate a possible deviation from the mean. For any $0 < t < 1$
\begin{equation}\label{eq:chernoff-ineq}
    \mathbb P(|Y_i - \E Y_i| > t \E Y_i) \leq e^{-\E Y_i t^2 /2}.
\end{equation}
Let us take $t = \frac{2 \sqrt{\log n}}{\sqrt{(n/2 - 1)\muin}}$. Then for large enough $n$,
\[
    \mathbb P\left(\left|\sum_{j = 1}^{n/2}{A_{ij}} - \muin \frac{n}{2} \right| > \sqrt{2\muin n\log n}\right)
    \weq
    \mathbb P\left(|Y_i - \E Y_i| > \sqrt{2\muin n\log n}\right)
    \wle
    \frac 1{n^2}.
\]
Similarly,
\[
    \mathbb P\left(\left|\sum_{j = n/2 + 1}^{n}{A_{ij}} - \muout \frac{n}{2} \right| > \sqrt{2\muout n\log n}\right)
    \wle
    \frac 1{n^2}
\]
and
\begin{equation}\label{y_i}
    \mathbb P\left( \left|u_i - (\muin - \muout) \frac{\sqrt n}{2} \right| > \sqrt{2(\muin + \muout) \log n}\right)
    \wle
    \frac 2{n^2}.
\end{equation}
Denote $\gamma_n = \sqrt{2(\muin + \muout) \log n}$ and notice that $\gamma_n = \Theta(\sqrt{\log n})$. By the union bound, we have for large enough $n$
\begin{equation}\label{u_i_a_b}
    \mathbb P\left(\exists i \leq \frac{n}{2} \colon \left|u_i - (\muin - \muout) \frac{\sqrt n}{ 2 }\right| > \gamma_n \right)
    \wle
    \frac{n}{2} \cdot \frac{2}{n^2}
    \weq
    \frac{1}{n}.
\end{equation}
By the same argumentation,
\begin{equation}\label{u_i_b_a}
    \mathbb P\left(\exists i > \frac{n}{2} \colon \left|u_i - (\muout - \muin) \frac{\sqrt n}{2}\right| > \gamma_n \right) \leq \frac1{n}.
\end{equation}
Now let us calculate $\rho(v_*)$:
\[
    \rho(v_*)
    \weq
    \sum_{i = 1}^{n} (v_*)_i u_i
    \weq
    \frac 1{\sqrt n} \sum_{i = 1}^{n/2}  u_i - \frac 1{\sqrt n} \sum_{i = n/2 + 1}^{n}  u_i.
\]
We already established that $u_i \sim (\muin - \muout) \frac{\sqrt n}{2}$ for $i \leq \frac{n}{2}$ (which means that $\lim \frac{2 u_i}{(\mu_{in} - \mu_{out})\sqrt n} = 1$ w.h.p.) and, therefore, that $\frac {1}{\sqrt n} \sum_{i = 1}^{n/2}  u_i \sim (\muin - \muout) \frac{n}{4}$. More precisely, by \eqref{u_i_a_b},
\[
    \mathbb P\left( \left| \frac 1{\sqrt n} \sum_{i = 1}^{n/2}  u_i - \frac{(\muin - \muout) n}4 \right| > \frac{\gamma_n \sqrt n} 2 \right)
    \wle
    \frac 1{n}.
\]
In the same way, by \eqref{u_i_b_a},
\[
    \mathbb P\left( \left| \frac 1{\sqrt n} \sum_{i = \frac{n}{2} + 1}^{n}  u_i - \frac{(\muout - \muin) n}4 \right| > \frac{\gamma_n \sqrt n} 2 \right)
    \wle
    \frac 1{n}.
\]
Finally,
\begin{equation}\label{rho-bound}
    \mathbb P\left( \left|\rho(v_*) - \frac{(\muin - \muout) n}2 \right| > \gamma_n \sqrt n \right)
    \wle
    \frac 2{n}.
\end{equation}

Now let us denote $w = Av_* - \rho(v_*) v_* = u - \rho(v_*) v_*$. As we already know, $u_i \sim (\muin - \muout) \frac{\sqrt n}{2}$ and $(\rho(v_*) v_*)_i \sim (\muin - \muout) \frac{\sqrt n}{2}$  for $i \leq \frac{n}{2}$. Clearly, for $i \leq \frac{n}{2}$
\[
    |w_i|
    \wle
    \left|u_i - \frac{(\muin - \muout)\sqrt n}2 \right| + \left|\frac{(\muin - \muout)\sqrt n}2 - \frac 1{\sqrt n}\rho(v_*) \right|.
\]
Then
\begin{align*}
    \mathbb P\left(|w_i| > \gamma_n \right) & \wle \mathbb P\left(\left|u_i - \frac{(\muin - \muout)\sqrt n}2 \right| > \gamma_n \right) + \\
    & \quad + \mathbb P\left(\left| \frac{(\muin - \muout)\sqrt n}2 - \frac 1{\sqrt n} \rho(v_*) \right| > \gamma_n \right).
\end{align*}
A similar bound can be derived for the case $i > n/2$. Taking into account that $\rho(v_*)$ does not depend on $i$, using the union bound and equations~\eqref{y_i} and~\eqref{rho-bound}, we get that
\[
    \mathbb P\left(\max_i |w_i| > 2\gamma_n \right)
    \wle
    n \cdot \frac 2{n^2} + \frac 2{n} \weq  \frac 4{n}.
\]
One can readily see that $\|w\|_2 \leq \sqrt{n \cdot \max_i w^2_i} = \sqrt n  \max_i |w_i|$. Thus, we finally can bound the Euclidean norm of the vector $w$:
\[
    \mathbb P \left( \|w\|_2 > 2\gamma_n \sqrt n \right)
    \wle \frac 4n \to 0, \;\; n\to\infty.
\]

Now we can use Theorem~\ref{th-kanan-parlett}. According to this result,
\[
    |\sin \angle(v_*, \widetilde v)| \wle
    \frac{\| A v_* - v_*\rho(v_*) \|_2}{\|v_*\|_2 \delta}
    \weq \frac{\|w\|_2}{\delta}
    \wle \frac{2 \gamma_n \sqrt n}{\delta} \;\; w.\,h.\,p.,
\]
where $\delta = \min_i |\lambda_i(A) - \rho(v_*)|$ over all $\lambda_i \neq \widetilde \lambda$. Since we have assumed that \eqref{condition:Fourier_mode_different_ideal_mode_meansum} and \eqref{condition:Fourier_mode_different_ideal_mode_meandifference} hold, by Proposition~\ref{proposition:unicity_eigenvector_general_case}, $\delta > \epsilon n$. Then, since $v_*$ is normalized, a simple geometric consideration guarantees that
\begin{equation}\label{ineq:norm_v_star_v_hat}
    \|v_* - \widetilde v\|_2
    \wle
    \sqrt 2 \, |\sin \angle(v_*, \widetilde v)|
    \wle
    \frac{2\sqrt 2\gamma_n \sqrt n}{\epsilon n}
    \weq \frac{2\sqrt 2\gamma_n}{\epsilon \sqrt n} \;\; w.\,h.\,p.
\end{equation}

Let us denote {\it the number of errors} by
\[
    r \weq \left|\left\{ i \in [n] \colon \sign\left( (v_*)_i \right) \neq \sign\left( \widetilde v_i \right) \right\} \right|.
\]
If we now remember that the vector $v_*$ consists of $\pm \frac{1}{\sqrt n}$, it is clear that for any $i$ with $\sign((v_*)_i) \neq \sign(\widetilde v_i)$
\[
    \left| (v_*)_i - \hat v_i \right| \ > \ \frac 1{\sqrt n}.
\]
The number of such coordinates is $r$. Therefore,
\[
    \|v_* - \widetilde v \|_2^2 \wge r \left( \frac 1{\sqrt n}\right)^2 \weq \frac rn.
\]
Then, by \eqref{ineq:norm_v_star_v_hat}, the following chain of inequalities holds:
\[
    \frac rn \wle \|v_* - \widetilde v \|_2^2
    \wle \frac{8\gamma_n^2}{\epsilon^2 n} \weq \frac{16 (\muin + \muout)\log n}{\epsilon^2 n} \;\; w.\,h.\,p.
\]
Hence, with high probability
\[
    r
    \wle \frac{16 (\muin + \muout)\log n}{\epsilon^2} = O(\log n), \;\; n\to\infty.
\]
Thus, the vector $\widetilde v$ provides almost exact recovery. This completes the proof.
\end{proof}

\subsection{Strong consistency of higher-order spectral clustering with local improvement}

In order to derive a strong consistency result, we shall add an extra step to Algorithm~\ref{algo:higher_order_sc}. Given $\widetilde{\sigma}$, the prediction of Algorithm~\ref{algo:higher_order_sc}, we classify each node to be in the community where it has the most neighbors, according to the labeling $\widetilde{\sigma}$. We summarize this procedure in Algorithm~\ref{algo:higher_order_sc_local_improvement}, and Theorem~\ref{thm:exact_recovery_sgbm} states the exact recovery result.

\begin{algorithm}
	\KwInput{Adjacency matrix $A$, average intra- and inter-cluster edge densities $\muin, \muout$.
	}
	
	\KwOutput{Node labelling $\widehat{\sigma} \in \{1,2\}^n$.}~\\
	\KwGlobalStep \\
	{   Let $\widetilde{\sigma}$ be the output of Algorithm~\ref{algo:higher_order_sc}.
	}~\\
	\KwLocalStep \\
	\For{$i=1,\dots,n$}
	{
	    Set $\hsigma_i := \argmax\limits_{k \in \{1,2\} } \sum\limits_{ j \not= i} 1\left(   \widetilde{\sigma}_j  = k \right) a_{ij}  $.
	}
	\caption{Higher-Order Spectral Clustering with Local Improvement (HOSC-LI).}
	\label{algo:higher_order_sc_local_improvement}
\end{algorithm}

\begin{remark}
    The local improvement step runs in $O(n d_{\rm max})$ operations, where $d_{\rm max}$ is the maximum degree of the graph.
    Albeit the local improvement step being convenient for the theoretical proof, we shall see in Section~\ref{section:numerical_results} (Figure~\ref{fig:accuracy_function_of_N}) that in practice Algorithm~\ref{algo:higher_order_sc} already works well, often giving 100\% accuracy even without local improvement.
\end{remark}

\begin{theorem}\label{thm:exact_recovery_sgbm}
    Let us consider the $d$-dimensional SGBM defined by~\eqref{eq:def_euclidean_block_model}-\eqref{eq:def_homogeneous_model}, and connectivity probability functions $\Fin$ and $\Fout$ satisfying conditions~\eqref{condition:Fourier_mode_different_ideal_mode_meansum}-\eqref{condition:Fourier_mode_different_ideal_mode_meandifference}. Then Algorithm \ref{algo:higher_order_sc_local_improvement} provides exact recovery for the given SGBM.
\end{theorem}

\begin{proof}
    We need to prove that the almost exact recovery of Algorithm~\ref{algo:higher_order_sc} (established in Theorem~\ref{thm:almost_recovery_sgbm}) can be transformed into exact recovery by the local improvement step. This step consists in counting neighbours in the obtained communities. For each node $i$ we count the number of neighbours in both supposed communities determined by the sign of the vector $\widetilde v$ coordinate:
\begin{align*}
    \widetilde Z_1(i) & \weq \sum_{\sign\left( \widetilde v_j \right) = 1} A_{ij};\\
    \widetilde Z_2(i) & \weq \sum_{\sign \left( \widetilde v_j \right) = -1} A_{ij}.
\end{align*}
According to Algorithm \ref{algo:higher_order_sc_local_improvement}, if $\widetilde Z_1(i) > \widetilde Z_2(i)$, we assign the label $\widehat\sigma_i = 1$ to node~$i$, otherwise we label it as $\widehat\sigma_i = 2$. Suppose that some node $i$ is still misclassified after this procedure and our prediction does not coincide with the true label: $\widehat\sigma_i \neq \sigma_i$. Let us assume without loss of generality that $\sigma_i = 1$ and, therefore, $\widehat \sigma_i = 2$. Then, clearly, $\widetilde Z_2(i) > \widetilde Z_1(i)$.

Let us denote by $Z_1(i)$ and $Z_2(i)$ degrees of node $i$ in the communities defined by the true labels~$\sigma$:
\[
    Z_j(i) \weq \sum_{\sigma_i = j} A_{ij}, \quad j = 1,2.
\]
Since $\sign(\widetilde v_j)$ coincides with the true community partition for all but $C\log n$ nodes (see the end of the proof of Theorem~\ref{thm:almost_recovery_sgbm}),
\[
        \left| \widetilde Z_j(i) - Z_j(i) \right| \wle C\log n, \quad j = 1,2,
\]
which implies that
\begin{align*}
        \widetilde Z_1(i) \wge Z_1(i) -  C\log n;\\
        \widetilde Z_2(i) \wle Z_2(i) + C\log n.
\end{align*}
Hence, taking into account that $\widetilde Z_2(i) \, > \, \widetilde Z_1(i)$,
\[
    Z_2(i) + 2C\log n \ > \ Z_1(i),
\]
which means that the inter-cluster degree of node $i$ is asymptotically not less than its intra-cluster degree (since $Z_j(i) = \Theta(n) \;\;  w. h. p.$). Intuitively, this should happen very seldom, and Lemma \ref{lem: neighb-counting} in the Appendix gives an upper bound on the probability of this event. Thus, by Lemma \ref{lem: neighb-counting}, for large $n$,
\begin{align*}
    &\mathbb P(Z_2(i) + 2C\log n > Z_1(i))
    \weq \mathbb P(Z_1(i) - Z_2(i) < 2C\log n) \wle \\
    & \wle \mathbb P\left(Z_1(i) - Z_2(i) < \sqrt{2(\muin + \muout)n\log n}\right) \wle \frac 1n \to 0, \;\; n\to\infty.
\end{align*}
Then each node is classified correctly with high probability and Theorem~\ref{thm:exact_recovery_sgbm} is proved.
\end{proof}

\section{Numerical results}
\label{section:numerical_results}

\subsection{Higher-order spectral clustering on 1-dimensional GBM}

Let us consider a 1-dimensional GBM, defined in Example~\ref{example:GBM-d_dimension}.
We first emphasize two important points of the theoretical study: the ideal eigenvector for clustering is not necesarily the Fiedler vector, and some eigenvectors, including the Fiedler vector, could correspond to geometric configurations.

\begin{figure}[!ht]
		\centering
		\includegraphics[width=0.45\textwidth]{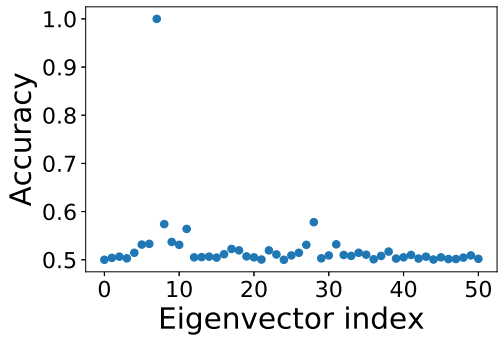}
		\caption{Accuracy obtained on a 1-dimensional GBM ($n = 2000$, $\rin =0.08$, $\rout = 0.02$) using the different eigenvectors of the adjacency matrix. The eigenvector of index $k$ corresponds to the eigenvector associated with the $k$-th largest eigenvalue of $A$.
		}
		\label{fig:accuracy_per_eigenvector}
    \end{figure}

Figure~\ref{fig:accuracy_per_eigenvector} shows the accuracy (\textit{i.e.,} the ratio of correctly labeled nodes, up to a global permutation of the labels if needed, divided by the total number of nodes) of each eigenvector for a realization of a 1-dimensional GBM. We see that, although the Fiedler vector is not suitable for clustering, there is nonetheless one eigevector that stands out of the crowd.

\begin{figure}[!ht]
	\centering
	\begin{subfigure}[b]{0.32\textwidth}
		\centering
		\includegraphics[width=\textwidth]{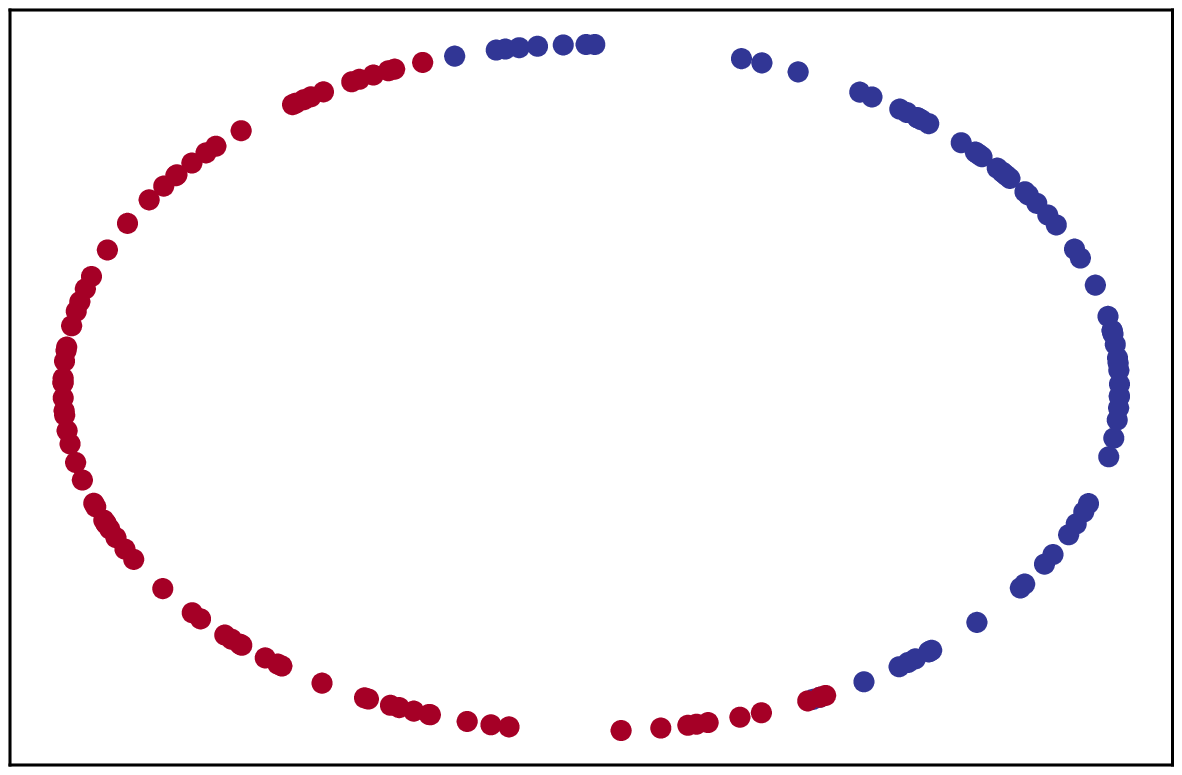}
		\caption{ $k = 2$ }
		\label{fig:eigenvector_prediction_k2}
	\end{subfigure}
	\hfill
	\begin{subfigure}[b]{0.32\textwidth}
		\centering
		\includegraphics[width=\textwidth]{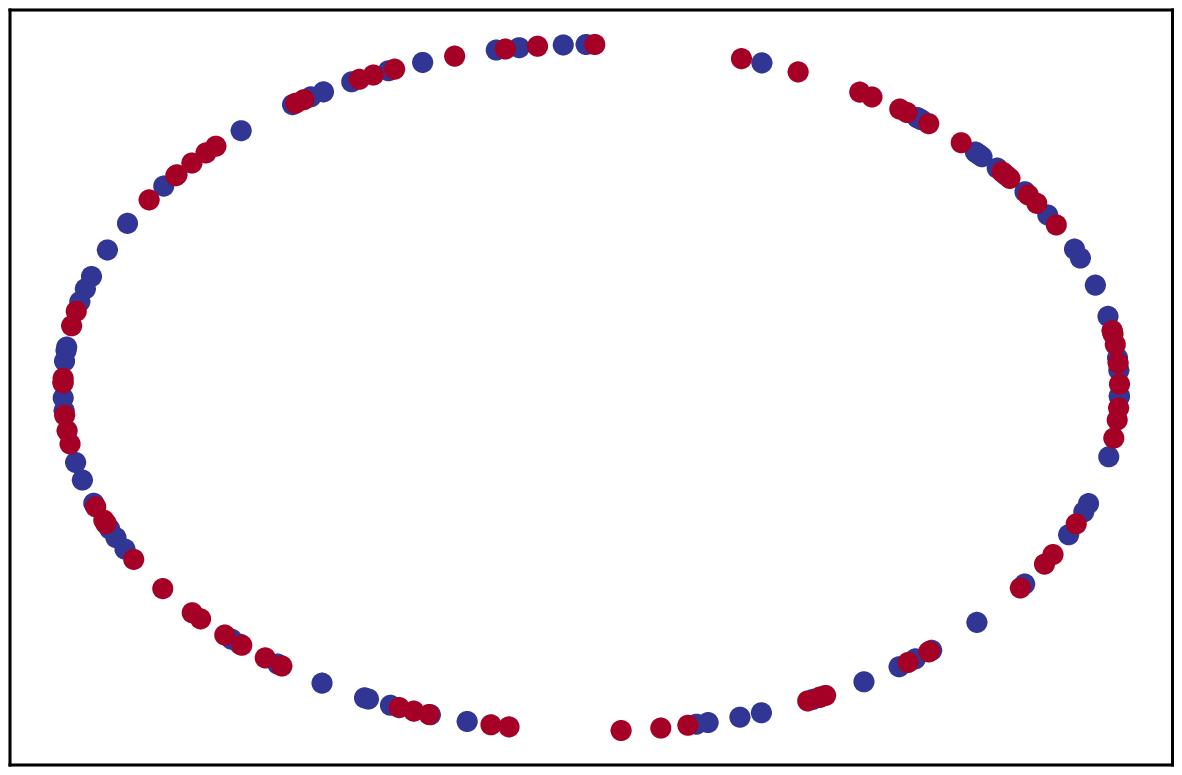}
		\caption{ $k = 4$}
		\label{fig:eigenvector_prediction_k4}
	\end{subfigure}
	\hfill
	\begin{subfigure}[b]{0.32\textwidth}
		\centering
		\includegraphics[width=\textwidth]{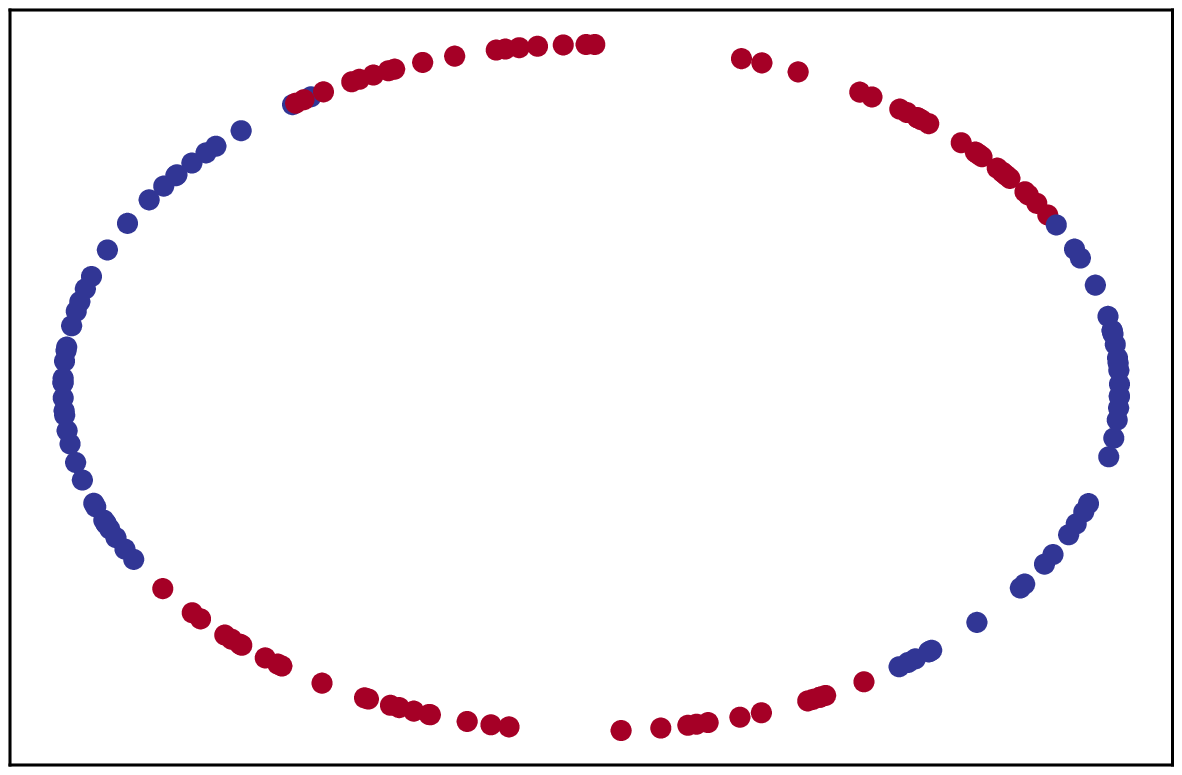}
		\caption{  $k = 8$ }
		\label{fig:eigenvector_prediction_k8}
	\end{subfigure}
	\caption{ (Best viewed in colour)
	Example of clustering done using the eigenvector associated to the $k$-th largest eigenvalue of the adjacency matrix of a 1-dimensional GBM ($n = 150$, $\rin = 0.2$, $\rout = 0.05$).
	For clarity edges are not shown, and nodes are positioned on a circle according to their true positions.
	The Fiedler vector ($k = 2$) is associated with a geometric cut, while the 4-th eigenvector corresponds to the true community labelling and leads to the perfect accuracy. The vector $k = 8$ is associated with yet another geometric cut.
	}
	\label{fig:eigenvector_prediction}
\end{figure}

Then, in Figure~\ref{fig:eigenvector_prediction} we draw the nodes of a GBM according to their respective position. We then show the clusters predicted by some eigenvectors. We see some geometric configurations (Figures~\ref{fig:eigenvector_prediction_k2} and~\ref{fig:eigenvector_prediction_k8}), while the eigenvector leading to the perfect accuracy corresponds to index 4 (Figure~\ref{fig:eigenvector_prediction_k4}).

\begin{figure}[!ht]
		\centering
		\includegraphics[width=0.45\textwidth]{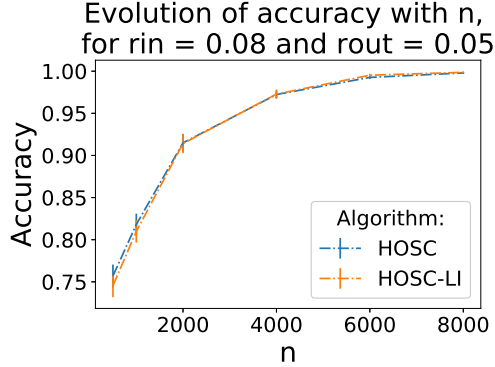}
		\caption{Accuracy obtained on $1$-dimensional GBM as a function of $n$, when $\rin = 0.08$ and $\rout = 0.05$, for Algorithm~\ref{algo:higher_order_sc} and Algorithm~\ref{algo:higher_order_sc_local_improvement}.
		Results are averaged over 100 realisations, and error bars show the standard error.
		}
		\label{fig:accuracy_function_of_N}
	\end{figure}

Figure~\ref{fig:accuracy_function_of_N} shows the evolution of the accuracy of Algorithms~\ref{algo:higher_order_sc} and~\ref{algo:higher_order_sc_local_improvement} when the number of nodes $n$ increases. As expected, the accuracy increases with $n$.
Moreover, we see no significant effect of using the local improvement of Algorithm~\ref{algo:higher_order_sc_local_improvement}. Thus, we conduct all the rest of numerical experiments with the basic Algorithm~\ref{algo:higher_order_sc}.

\begin{figure}[!ht]
    \centering
    \includegraphics[width=0.45\linewidth]{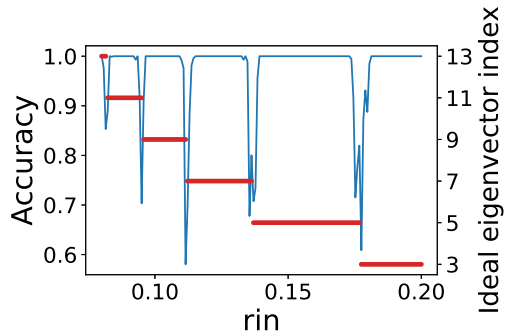}
    \caption{(Best viewed in colour) Evolution of accuracy (blue curve) with respect to  $\rin$, for a GBM with $n = 3000$ and $\rout = 0.06$. Results are averaged over 5 realisations. By the red curve we show the index of the ideal eigenvector, again with respect to $\rin$.}
    \label{fig:accuracy_dip_bad_values}
\end{figure}

In Figure~\ref{fig:accuracy_dip_bad_values}, we illustrate the statement of Proposition~\ref{cor:gap_between_good_eigenvaue_and_others}: for some specific values of the pair $(\rin, \rout)$, the Conditions~\eqref{condition:Fourier_mode_different_ideal_mode_meansum} and~\eqref{condition:Fourier_mode_different_ideal_mode_meandifference} do not hold, and Algorithm~\ref{algo:higher_order_sc} is not guaranteed to work. We observe in Figure~\ref{fig:accuracy_dip_bad_values} that these pairs of bad values exactly correspond to the moments when the index of the ideal eigenvector jumps from one value to another.

\begin{figure}[!ht]
			\centering
			\includegraphics[width=0.45\textwidth]{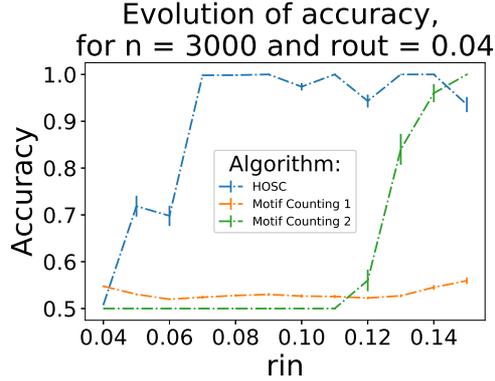}
		\caption{(Best viewed in colour) Accuracy obtained on $1$-dimensional GBM for different clustering methods. Motif Counting 1 corresponds to the algorithm described in \cite{Galhotra_al_2018} and Motif Counting 2 to the algorithm described in \cite{Galhotra_al_2018_2nd_paper}. Results are averaged over 50 realisations, and error bars show the standard error.
        }
		\label{fig:comparison_with_umass_algorithms}
	\end{figure}

Finally, we compare in Figure~\ref{fig:comparison_with_umass_algorithms} the accuracy of Algorithm~\ref{algo:higher_order_sc} with the motif counting algorithms presented in references~\cite{Galhotra_al_2018} and \cite{Galhotra_al_2018_2nd_paper}. Those algorithms consist in counting the number of common neighbors, and clustering accordingly. We call Motif Counting 1 (resp. Motif Counting 2) the algorithm of reference~\cite{Galhotra_al_2018} (resp. of reference~\cite{Galhotra_al_2018_2nd_paper}).
We thank the authors for providing us the code used in their papers.
We observed that with present realizations the motif counting algorithms take much more time than HOSC takes.
For example on a GBM with $n = 3000$, $\rin = 0.08$ and $\rout = 0.04$, HOSC takes 8 seconds, while Motif Counting~1 takes 130 seconds and Motif Counting~2 takes 60 seconds on a laptop with 1.90GHz CPU and 15.5 GB memory.

\subsection{Waxman Block Model}

    Let us now consider the Waxman Block Model introduced in Example~\ref{example:Waxman_Block_Model}.
    Recall that $\Fin(x) = \min(1, q_{\rm in} e^{-s_{\rm in} x})$ and $\Fout(x) = \min(1, q_{\rm out} e^{-s_{\rm out} x})$, where $q_{\rm in}, q_{\rm out}, s_{\rm in}, s_{\rm out}$ are four positive real numbers.
    We have the following particular situations:
    \begin{itemize}
        \item if $\sout = 0$, then $\Fout (x) = \qout$ and the inter-cluster interactions are independent of the nodes' positions. If $s_{\rm in} = 0$ as well, we recover the SBM;
        \item if $\qin = e^{\rin s_{\rm in} }$ and $\qout = e^{\rout \sout}$, then in the limit $s_{\rm in}, \sout \gg 1$ we recover the 1-dimensional GBM.
    \end{itemize}

\begin{figure}[!ht]
		\begin{subfigure}[b]{0.47\textwidth}
			\centering
			\includegraphics[width=\textwidth]{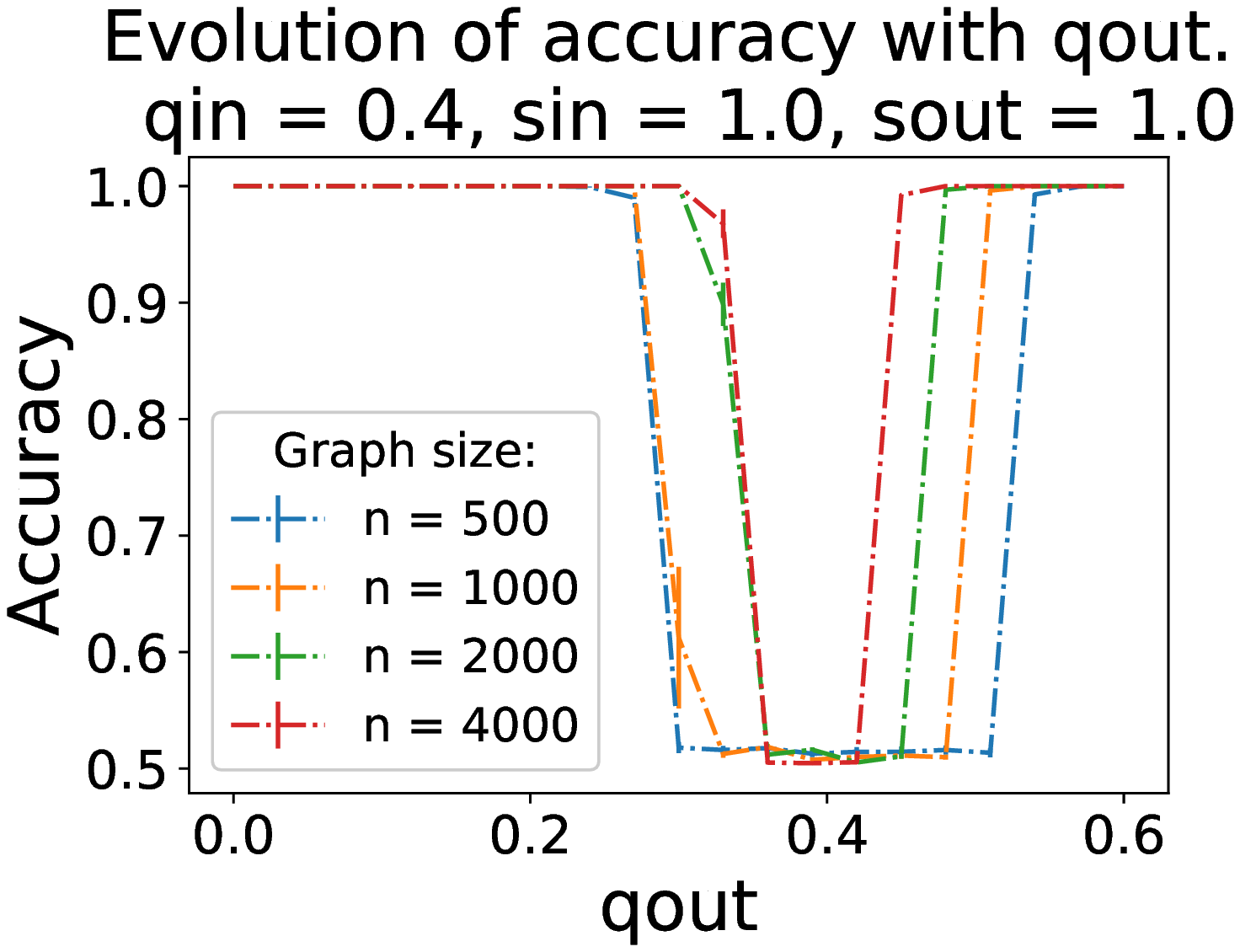}
		\end{subfigure}
		\hfill
		\begin{subfigure}[b]{0.47\textwidth}
			\centering
			\includegraphics[width=\textwidth]{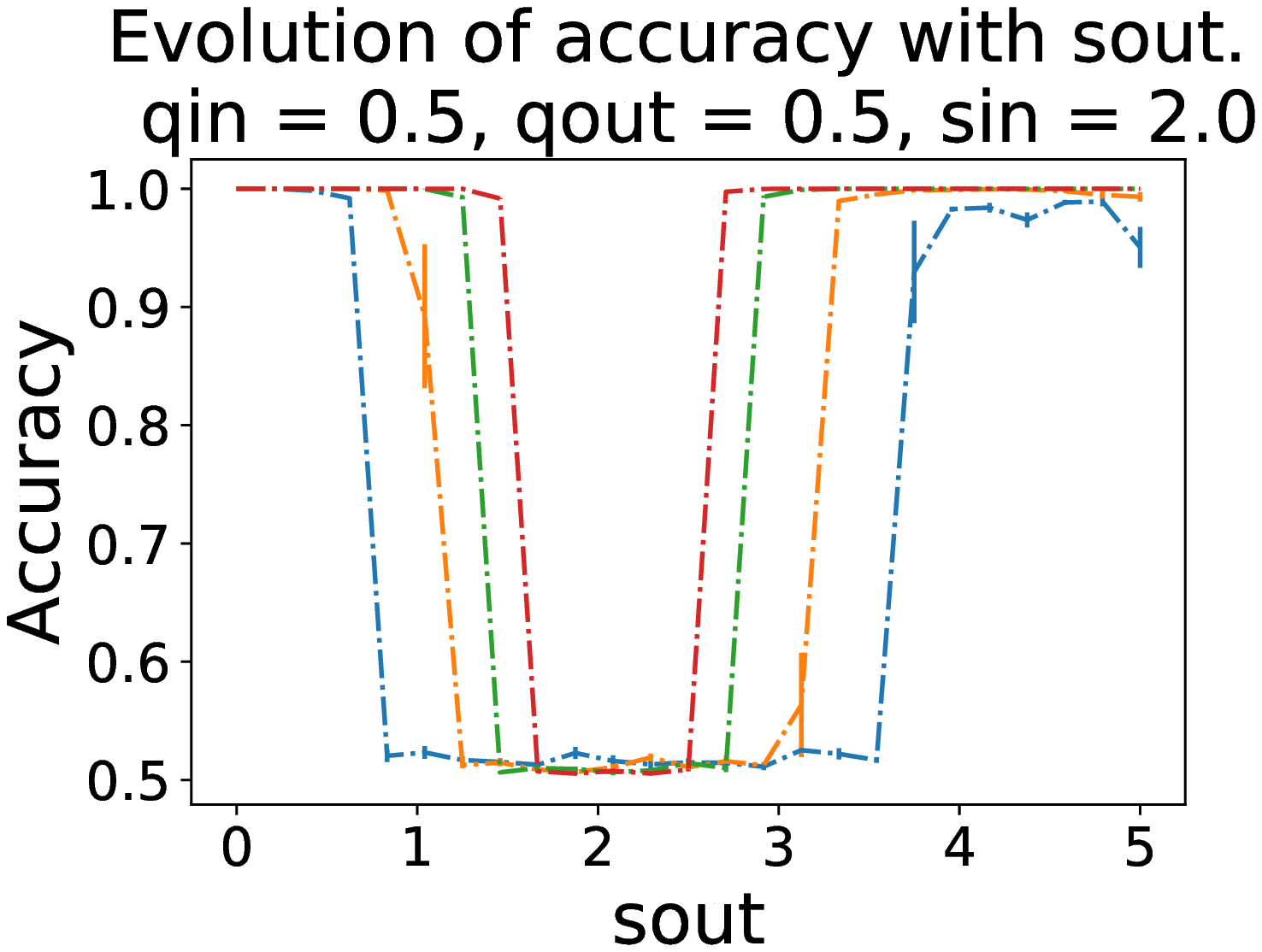}
		\end{subfigure}
		\caption{(Best viewed in colour) Accuracy obtained on a 1-dimensional Waxman Block Model.
		Results are averaged over 10 realizations. Same colors in the two plots correspond to the same graph size.
		}
		\label{fig:accuracy_waxman}
\end{figure}

    We show in Figure~\ref{fig:accuracy_waxman} the accuracy of Algorithm~\ref{algo:higher_order_sc} on a WBM. In particular, we see that we do not need $\muin> \muout$, and we can recover disassociative communities.
    However, there are obvious dips when $\qin$ is close to $\qout$ or $s_{\rm in}$ is close to $\sout$. It is clear that if $\qin = \qout$ on the left-hand side picture or $s_{\rm in} = \sout$ on the right-hand side picture, one cannot distinguish two communities in the graph. Thus, for small $n$, we observe some ranges around these `bad' values where  Algorithm~\ref{algo:higher_order_sc} fails. As expected, the dips become narrower when~$n$ increases.

\section{Conclusions and future research}
\label{section:conclusion}

In the present paper we devised an effective algorithm for clustering geometric graphs. This algorithm is close in concept to the classical spectral clustering method but it makes use of the eigenvector associated with a higher-order eigenvalue. It provides weak consistency for a wide class of graph models which we call the Soft Geometric Block Model, under some mild conditions on the Fourier transform of $\Fin$ and $\Fout$. A small adjustment of the algorithm leads to strong consistency. Moreover, our method was shown to be effective in numerical simulations even for graphs of modest size.

The problem stated in the current paper might be investigated further in several directions. One of them is a possible study on the SGBM with more than two clusters. The situation here is quite different from the SBM where the spectral clustering algorithm with few eigenvectors associated with the smallest non-zero eigenvalues provides good performance. In the SGBM even the choice of such eigenvectors is not trivial, since the `optimal' eigenvalue for distinguishing two clusters is often not the smallest one.

Another natural direction of research is the investigation of the sparse regime, since all our theoretical results concern the case of degrees linear in $n$ (this assumption is used for the analysis of the adjacency matrix spectrum and for finding the spectral gap around the `ideal' eigenvalue $\tilde \lambda$). In sparser regimes, there are effective algorithms for some particular cases of the SGBM (e.\,g., for the GBM), but there is no established threshold when exact recovery is theoretically possible. Unfortunately, the method of the current paper without revision is not appropriate for this situation, and the technique will very likely be much more complicated.

Finally, considering weighted geometric graphs could be an interesting task for applications where clustering of weighted graphs is pertinent. For instance, the functions $\Fin$ and $\Fout$ can be considered as weights on the edges in a graph. We believe that most of the results of the paper may be easily transferred to this case.

\section*{Acknowledgements}
This work has been done within the project of Inria - Nokia Bell Labs ``Distributed Learning and Control for Network Analysis''.

The authors are most grateful to
Laura Cottatellucci, Mounia Hamidouche and Mindaugas Bloznelis for discussions that helped to stimulate this research.

\bibliographystyle{plain}
\bibliography{hosc_arxiv}   % name your BibTeX data base

\appendix
\section{Auxiliary results}

\subsection{Hamburger moment problem for the limiting measure}

\begin{lemma}\label{lemma:moment_problem_mu} Assume that $F: \T^d \to \mathbb R$ is such that $F(0)$ is equal to the Fourier series of $F(x)$ evaluated at~$0$ and $0 \leq F(x) \leq 1$. Consider the measure $\mu$ defined by the function $F$:
$$
    \mu = \sum_{k \in \mathbb Z^d} \delta_{\widehat F(k)}.
$$
Then $\mu$ is defined uniquely by the sequence of its moments $\{M_n\}_{n = 1}^{\infty}$.
\end{lemma}

\begin{proof}
It is enough to show that Carleman's condition holds true for $\mu$ (see \cite{aheizer_1965}):
\begin{equation}\label{eq:carleman_condition}
    \sum_{n = 1}^{\infty} M_{2n}^{-\frac 1{2n}} = +\infty.
\end{equation}
As one can easily see,
\begin{equation}\label{eq:moment_expression}
    M_{2n} = \sum_{k\in \mathbb Z^d} \widehat F^{2n}(k).
\end{equation}

From the bounds $0  \leq F(x) \leq 1$ it follows that $\widehat F(k) \leq 1$. Then it is clear that $\widehat F^n(k) \leq \widehat F(k)$ for any $n \in \mathbb N$. We can write
$$
    M_{2n} = \sum_{k\in \mathbb Z^d} \widehat F^{2n}(k) \leq \sum_{k\in \mathbb Z^d} \widehat F(k) = F(0) \leq 1.
$$
Here we used the assumption that $F(0)$ equals its Fourrier series evaluated at $0$. Then
$$
    M_{2n}^{-\frac 1{2n}} \geq 1,
$$
Thus, the series in the right-hand side of \eqref{eq:carleman_condition} is divergent and Carleman's condition is verified.

\end{proof}

\subsection{\texorpdfstring{$m$}{m}-times convolution}

\begin{lemma}\label{appendix:lemma_convolution_m_times}
    Let $m\in \N$ and $F_1, \dots, F_m$ be integrable functions over $\T^d$. Then,
        \[
        F_1 * \dots * F_m (0) = \int_{(\T^d)^m} \prod_{j=1}^m F_j(z_j - z_{j+1}) \ dz_1 \dots dz_m
        \]
        with the notation $z_{m+1} = z_1$.
\end{lemma}

\begin{proof}
    With the change of variable $u_i = z_i - z_{i+1}$ for $i=1,\dots, m-1$, we have
    \begin{align*}
        & \int_{(\T^d)^m} \prod_{j=1}^m F_j(z_j - z_{j+1}) \ dz_1 \dots d  z_m \\ \weq & \int_{\T^d} dz_1 \int_{ (\T^d)^{m-1} } \prod_{i=1}^{m-1} F_i (u_i) F_m(-u_1 - \dots - u_{m-1}) du_1 \dots du_{m-1}
    \end{align*}
    We notice that
    \begin{align*}
        & \int_{\T^d} du_{m-1} F_{m-1}(u_{m-1}) F_m(-u_1-\dots - u_{m-1} )  \\ \weq & F_{m-1} * F_m (-u_1 - \dots - u_{m-2}).
    \end{align*}
    Hence,
    \begin{align*}
         & \int_{ (\T^d)^{m-1} } \prod_{i=1}^{m-1} F_i (u_i) F_m(-u_1 - \dots - u_{m-1}) du_1 \dots du_{m-1} \\
         \weq & F_1 * \dots * F_m (0).
    \end{align*}
    Therefore,
    \begin{align*}
         \int_{(\T^d)^m} \prod_{j=1}^m F_j(z_j - z_{j+1}) \ dz_1 \dots dz_m
        & \weq \int_{\T^d} d z_1 F_1 * \dots * F_m (0) \\
        & \weq  F_1 * \dots * F_m (0),
    \end{align*}
    which ends the proof.
\end{proof}

\subsection{Fourier transform of the square wave}

\begin{lemma}\label{lemma:computation_fourier_transform_square_signal}Let $0 < r < \frac12$.
Let $F : \R^d \to \R$ be 1-periodic such that $F(x) = 1\left( \|x\| \leq r \right)$ for $x \in \T^d$. Then,
\[
\hF(k) \weq 2r^d \prod_{j = 1}^d \sinc(2\pi k_j r),
\]
where $k = (k_1, \ldots, k_d) \in \mathbb Z^d$ and
$$
\sinc (x) =
\begin{cases}
1, & \text{ if } x=0, \\
\frac{\sin x}{x}, & \text{ otherwise.}
\end{cases}
$$
\end{lemma}

\begin{proof}
We shall use the set $[-1/2, 1/2]^d$ as a representation of $\T^d$. Let us first notice that for $x \in [-1/2, 1/2]^d$
\[
    F(x) = 1(\|x\| \leq r) \weq 1\left( \max_{1 \leq j \leq d}|x_j| \, \leq \, r \right) \weq \prod_{j=1}^d  1\left( |x_j| \, \leq \, r \right).
\]
Then
\begin{align*}
    \hF(k) & \weq \int_{ \left[ -\frac12, \frac12 \right]^d} F(x) e^{-2\pi i \langle k,x \rangle} dx \\
    & \weq \int_{ \left[ -\frac12, \frac12 \right]^d } \prod_{j=1}^d 1(|x_j| \leq r) e^{-2\pi i k_j x_j} dx_1 \ldots dx_d \\
    & \weq \prod_{j=1}^d \int_{-1/2}^{1/2} 1(|x_j| \leq r) e^{-2\pi i k_j x_j} dx_j.
\end{align*}

Let us consider some $1 \leq j \leq d$. If $k_j = 0$, we have $\int_{-1/2}^{1/2} 1(|x_j| \leq r) d x_j = \int_{-r}^r d x = 2r$. Moreover, for $k_j \not= 0$,
\begin{align*}
    \hF(k) & \weq  \int_{-1/2}^{1/2} 1(|x_j| \leq r) e^{-2 \pi i k_j x_j} d x_j \weq \int_{-r}^r e^{-2\pi i k_j x_j} d x_j \weq \dfrac{ e^{-2\pi i k_j r} - e^{2\pi i k_j r}  }{ - 2\pi i k_j} = \\
            & \weq \dfrac{\sin(2 \pi k_j r)}{\pi k_j} \weq 2r \dfrac{\sin(2 \pi k_j r)}{ 2 \pi k_j r} = 2r \sinc(2\pi k_j r).
\end{align*}
Hence,
\[
    \hF(k) \weq 2r^d \prod_{j = 1}^d \sinc(2\pi k_j r),
\]
as was needed.

\end{proof}

\subsection{Number of neighbours in different clusters}
	
\begin{lemma}\label{lem: neighb-counting}
    Let us consider the SGBM with connectivity probability functions $\Fin$ and $\Fout$ for which $\muin = \hFin(0) > \hFout(0) = \muout$. Denote by $Z_{\rm in}(i)$ (resp., $Z_{\rm out}(i)$) the `intra-cluster' (resp., `inter-cluster') degree of $i$:
    \begin{align*}
        Z_{\rm in}(i) & \weq \sum_{j \colon \sigma_j = \sigma_i} A_{ij};\\
        Z_{\rm out}(i) & \weq \sum_{j \colon \sigma_j \neq \sigma_i} A_{ij}.
    \end{align*}
    Denote $\cB_i := \left\{Z_{\rm in}(i) -  Z_{\rm out}(i) \, < \, \sqrt{2(\muin + \muout) n \log n} \right\}$. Then
    $$
         \pr \left( \cup_{i=1}^n \cB_i \right) \leq \frac 1n.
    $$
\end{lemma}

\begin{proof}
    Let us fix $i \in [n]$. Clearly, $Z_{\rm in}(i) \sim \mathrm{Bin}(\frac{n}{2} - 1, \muin)$ and $Z_{\rm out}(i) \sim \mathrm{Bin}(\frac{n}{2}, \muout)$. We again use Chernoff inequality \eqref{eq:chernoff-ineq}. By this bound, for $t = \frac{2\sqrt{\log n}}{\sqrt{(n/2-1)\muin}}$ and large enough $n$
    \[
        \mathbb P\left( \left| Z_{\rm in}(i) - \muin \frac{n}{2} \right| \, > \, \sqrt{2\muin n\log n}\right) \wle \frac 1{n^2}.
    \]
    By the same reason, $Z_{out}$ is well concentrated around its mean $\muout \frac{n}{2}$:
    \[
        \pr \left( \left| Z_{\rm out}(i) - \muout \frac{n}{2} \right| \, > \, \sqrt{2\muout n\log n}\right) \wle \frac 1{n^2}.
    \]
    Therefore, since $\muin > \muout$,
    \[
      \pr ( \cB_i ) = \pr \left(Z_{\rm in}(i) - Z_{\rm out}(i) \, < \, \sqrt{2(\muin + \muout)n\log n} \right)
      \wle \frac{1}{n^2}.
    \]
    By the union bound,
    \[
        \pr ( \cB ) \wle n \pr ( \cB_1 ) \wle \frac{1}{n},
    \]
    which proves the lemma.
\end{proof}

\subsection{Trace operator is Lipschitz}

    \begin{lemma}\label{lemma:appendix_Tr_1_lischitz}
        Let $A, \widetilde{A} \in \{0,1\}^{n \times n}$ be two adjacency matrices, and $m \geq 1$. Then,
        \[
        \left| \Tr A^{m} - \Tr \widetilde{A}^{m} \right| \wle m \, n^{m-2} \dHam\left( A, \widetilde{A} \right).
        \]
    \end{lemma}

    \begin{proof}
        %\rnote{(Max) Could it be sensibly simplified ? The point is to show tha $Q_m$ is Lipshitz w.r.t Hamming distance (or another distance would be simpler ?). Maybe a trick is to take two matrices that differs only in one entry?}
	Since $A$ and $\widetilde{A}$ are adjacency matrices of graphs without self-loops, we have $\Tr A = 0 = \Tr \widetilde{A}$. Hence $\left| \Tr A - \Tr \widetilde{A} \right| =0 \leq \frac{1}{n} \dHam \left( A,\widetilde A \right) $, and the statement holds for $m = 1$.
	%Lemma~\ref{lemma:appendix_Tr_1_lischitz} is proved.

    Let us now consider $m \geq 2$. We have
	\begin{align*}
	    \left| \Tr \left(A^{m-1}\right) - \Tr ( \widetilde{A}^{m-1} ) \right| & \weq \left|  \sum_{i_1, \dots, i_m} \left( \prod_{j=1}^m A_{i_j i_{j+1} } - \prod_{j=1}^m \widetilde{A}_{i_j, i_{j+1}}  \right) \right| \\
	    & \wle \sum_{i_1, \dots, i_m} \left| \prod_{j=1}^m A_{i_j i_{j+1} } - \prod_{j=1}^m \widetilde{A}_{i_j, i_{j+1}}  \right|,
	\end{align*}
    with the notation $i_{m+1} = i_1$.
	The quantity $\prod_{j=1}^m A_{i_j i_{j+1} }$ is equal to one if $A_{i_j i_{j+1} } = 1$ for all $j=1,\dots, m$, and equals zero otherwise.
	Hence, the difference $\left| \prod_{j=1}^m A_{i_j i_{j+1} } - \prod_{j=1}^m \widetilde{A}_{i_j, i_{j+1}}  \right|$ is non-zero and is equal to one in two scenarii:
	\begin{itemize}
	    \item[$\bullet$] $A_{i_j i_{j+1}}$ = 1 for all $j = 1, \dots, m$, while there is a $j'$ such that $\widetilde A_{i_{j'} i_{j'+1}} = 0$,
	    \item[$\bullet$] there is a $j'$ such that $A_{i_{j'} i_{j'+1}} = 0$ and $\widetilde A_{i_j i_{j+1}}$ = 1 for all $j = 1, \dots, m$.
	\end{itemize}
	Thus,
	\begin{multline*}
	    \left| \prod_{j=1}^m A_{i_j i_{j+1} } - \prod_{j=1}^m \widetilde{A}_{i_j, i_{j+1}}  \right| \weq 1\left( \forall j \; A_{i_j i_{j+1} } = 1  \right) 1 \left( \exists j' \colon \widetilde A_{i_{j'} i_{j'+1}} = 0 \right) + \\
	    + 1 \left( \exists j' \colon A_{i_{j'} i_{j'+1}} = 0 \right) 1\left( \forall j \; \widetilde A_{i_j i_{j+1} } = 1  \right).
	\end{multline*}
	   But,
	   \begin{align*}
	   1\left( \forall j \; A_{i_j i_{j+1} } = 1  \right) 1 \left( \exists j' \colon \widetilde A_{i_{j'} i_{j'+1}} = 0 \right)
	   &\wle \prod_{j=1}^m 1(A_{i_j i_{j+1}} = 1 )  \sum_{j=1}^m 1 \left( \widetilde A_{i_{j} i_{j+1}} = 0 \right) \\
	   & \wle \sum_{j=1}^m 1(A_{i_j i_{j+1}} = 1 )  1(\widetilde A_{i_j i_{j+1}} = 0 ).
	   \end{align*}
    Similarly,
    \begin{align*}
         1 \left( \exists j' \colon A_{i_{j'} i_{j'+1}} = 0 \right) 1\left( \forall j \; \widetilde A_{i_j i_{j+1} } = 1  \right) & \wle \sum_{j=1}^m 1(A_{i_j i_{j+1}} = 0 )  1(\widetilde A_{i_j i_{j+1}} = 1 ).
    \end{align*}
	Therefore,
	\begin{align*}
	    \left| \prod_{j=1}^m A_{i_j i_{j+1} } - \prod_{j=1}^m \widetilde{A}_{i_j, i_{j+1}}  \right| & \wle \sum_{j=1}^m 1\left( A_{i_j i_{j+1}} \not = \widetilde A_{i_j i_{j+1}} \right).
	    %& \wle \dHam(A, \widetilde A).
	    %& \wle m  1\left( A_{i_1 i_{2}} \not = A'_{i_1 i_{2}} \right).
	\end{align*}
	This leads to
	\begin{align*}
	    \sum_{i_1, \dots, i_m}  \left| \prod_{j=1}^m A_{i_j i_{j+1} } - \prod_{j=1}^m \widetilde{A}_{i_j, i_{j+1}}  \right|
	    & \wle \sum_{i_1, \dots, i_m} \sum_{j=1}^m 1\left( A_{i_j i_{j+1}} \not = \widetilde A_{i_j i_{j+1}} \right) \\
	    & \wle m \ n^{m-2} \, \dHam(A,\widetilde A),
	\end{align*}
	where the last line holds since for all $j = 1, \dots, m$ and $m\geq 2$
	\begin{align*}
	    \sum_{i_1, \dots, i_m} 1\left( A_{i_j i_{j+1}} \not = \widetilde A_{i_j i_{j+1}} \right) & \weq n^{m-2} \sum_{i_j, i_{j+1}} 1\left( A_{i_j i_{j+1}} \not = \widetilde A_{i_j i_{j+1}} \right) \\
	    & \weq n^{m-2} \dHam \left( A, \widetilde A \right).
	\end{align*}
    \end{proof}

\end{document}